\documentclass{article} 
\usepackage{iclr2027_conference,times}
\usepackage{graphicx}
\usepackage{wrapfig}
\usepackage{booktabs}
\usepackage{multirow}
\usepackage{colortbl}\usepackage{tabularx}

\usepackage{amsmath,amsfonts,bm}

\def\eqref#1{equation~\ref{#1}}

\def\1{\bm{1}}

\DeclareMathAlphabet{\mathsfit}{\encodingdefault}{\sfdefault}{m}{sl}
\SetMathAlphabet{\mathsfit}{bold}{\encodingdefault}{\sfdefault}{bx}{n}

\usepackage{hyperref}
\usepackage{url}

\title{PhotoGuider: Composition-Aware Cropping \\ and 360° View Framing with Explainable \\ Photographic Assessments}

\author{\normalfont
\textbf{Yuan-Kang Lee\textsuperscript{1,3}} \ \textbf{Kuan-Lin Chen\textsuperscript{2}} \ \textbf{Chia-Che Chang\textsuperscript{1}} \ \textbf{Yu-Lun Liu\textsuperscript{3}} \\
\textsuperscript{1}MediaTek Inc. \ \textsuperscript{2}National Taiwan University \\
\textsuperscript{3}Department of Computer Science, National Yang Ming Chiao Tung University
}

\iclrfinalcopy 
\begin{document}
\raggedbottom

\maketitle
\lhead{Preprint. Under review.}

\begin{figure}[!htbp]
    \centering
    \includegraphics[width=\linewidth]{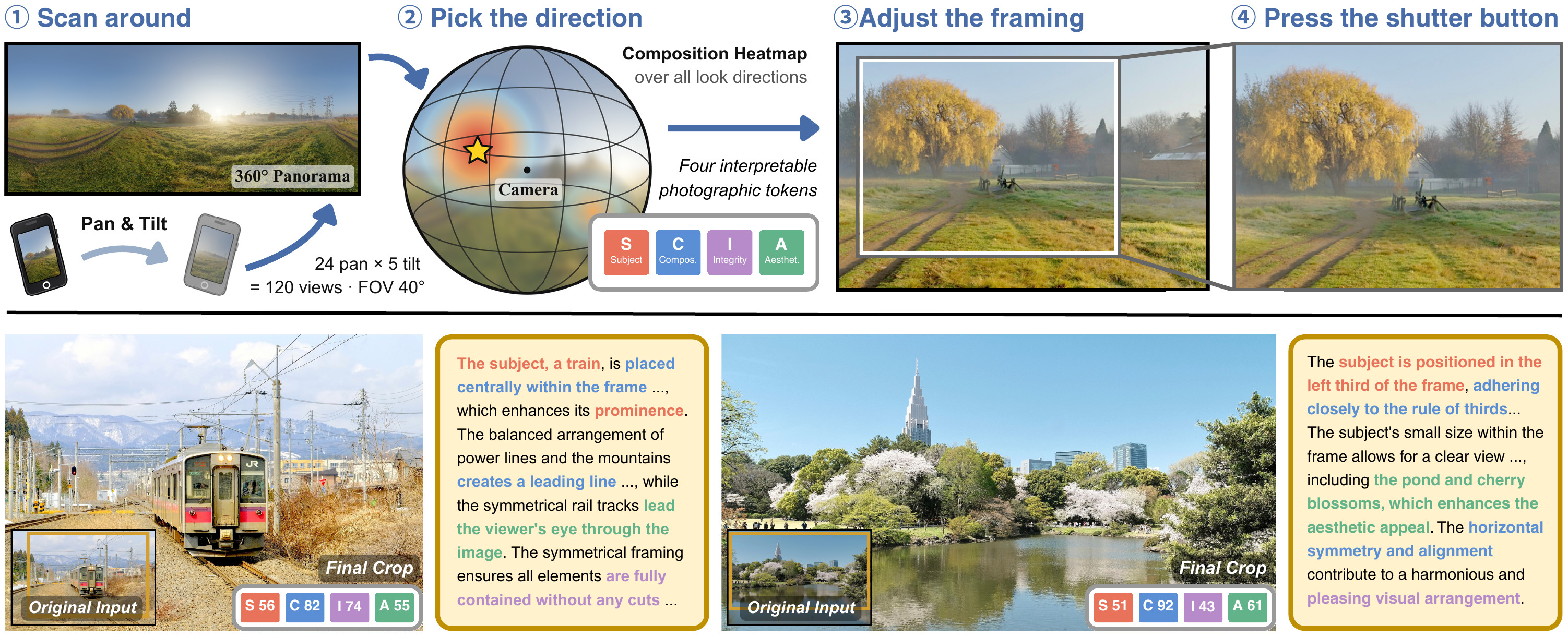}

    \vspace{-8pt}
    \caption{\textbf{PhotoGuider turns a whole scene into a framed photograph and provides a reasoned explanation for its choice of framing.} The four stages follow what a photographer actually does upon arriving at a scene: look around, settle on a direction, adjust the frame, and press the shutter. (1) Given a 360\textdegree{}~panorama, the system scans look directions and (2) scores each using the proposed four photographic tokens: Subject, Composition, Integrity, and Aesthetics. The maximum fixes the camera direction. (3) The frame is then refined within that view, and (4) the final photograph is rendered natively from the panorama. Our MLLM explainer is fine-tuned to cite these tokens and conditioned on their measured assessments. Below, two results show the selected crop and the produced explanation; phrases grounded in the four assessments are highlighted.}
    \label{fig:teaser}
\end{figure}

\begin{abstract}
Traditional image cropping assumes the photograph already contains the composition worth keeping, and asks only where to crop. A full 360\textdegree{}~scene offers no such guarantee: the viewing direction in a scene has to be discovered before anything can be framed. We present PhotoGuider, a measure-then-frame method that answers both questions with compact learned scorers over one shared set of composition-aware features. The designed features are extracted using foundation and task-specific vision models and structured into four photographic tokens: Subject, Composition, Aesthetics, and Integrity. Our MLLM explainer is fine-tuned to generate composition-related explanations conditioned on images and measured token assessments. To our knowledge, PhotoGuider is the first method to combine explicit composition-based direction selection in 360\textdegree{}~panoramas with refined framing, camera-parameter output for native re-rendering, and measurement-conditioned explanations. On two standard 2D cropping benchmarks, PhotoGuider achieves the highest average mIoU performance against ground-truth crops. In a blind ranking study, 120 participants ranked three cropping methods over 100 panoramic scenes from the public Poly Haven HDRI library, placing our framings first in 53.2\% of cases, against 27.3\% and 19.5\% for two strong alternatives. Project page: \url{https://ntuneillee.github.io/research/photoguider/}
\end{abstract}

\vspace{-2mm}
\section{Introduction}
\vspace{-2mm}

While computational photography has effectively matured low-level image acquisition~\citep{mohammadi2026low,brooks2019unprocessing,lee2024efficient,zhang2021rethinking,he2026blurdm,kong2026deblurdiff,feijoo2025darkir,kim2025ccmnet,lee2024learning,lee2026rl,lee2026safe}, higher-level compositional choices require separate treatment. When cropping an image, a photographer answers four fundamental questions: \textbf{\textit{what the subject is}}, \textbf{\textit{where to place it}}, \textbf{\textit{whether critical elements are cut off}}, and \textbf{\textit{if the overall composition is visually appealing}}. These four independent judgments are susceptible to distinct failures, such as selecting a correct subject with poor placement. Furthermore, panoramic framing introduces an earlier decision: determining direction before the frame is formed. Existing 2D cropping methods assume that the input is already a pre-shot photograph. Our goal is to combine composition-based direction selection with refined framing, camera-parameter output, and measurement-conditioned explanations. \textbf{PhotoGuider} evaluates candidate crops using four interpretable photographic tokens: Subject (S), Composition (C), Integrity (I), and Aesthetics (A). These assessments collectively score panoramic view directions, rank candidate frames, guide spatial refinements, and anchor MLLM explanations in concrete measurements. Pre-trained vision models extract the features, and compact scorers learn to combine them for framing decisions. This architecture offers three advantages. First, interpretability is structural at the feature level and post hoc at the attribution level. Second, the trainable decision layer has limited capacity to reduce the risk of overfitting to dataset-specific biases. Third, it natively generalizes to full panoramas, as the tokens evaluate intrinsic geometry and semantics on perspective projections. The overview of PhotoGuider is demonstrated in Fig.~\ref{fig:teaser}. Overall, PhotoGuider provides three contributions:

\begin{itemize}

\item Composition-Aware Cropping: The decision layer selects the optimal box using four photographic tokens and learned compact rankers with only 63k parameters, achieving competitive performance across two benchmarks while confining the MLLM to a bounded consultant role that neither scores nor makes decisions.

\item Measurement-Conditioned Explanations: We condition a fine-tuned MLLM on the token assessments and geometric measurements, allowing its measurement-based statements to be checked against the supplied values. 

\item To our knowledge, PhotoGuider is the first system for automatic composition-aware framing from 360\textdegree{}~panoramas that combines explicit view-level direction selection with camera-parameter output for re-rendering and measurement-conditioned explanations based on shared photographic feature groups.

\end{itemize}

\begin{figure}[!t]
\vspace{-1mm}
    \centering
    \includegraphics[width=1.00\linewidth]{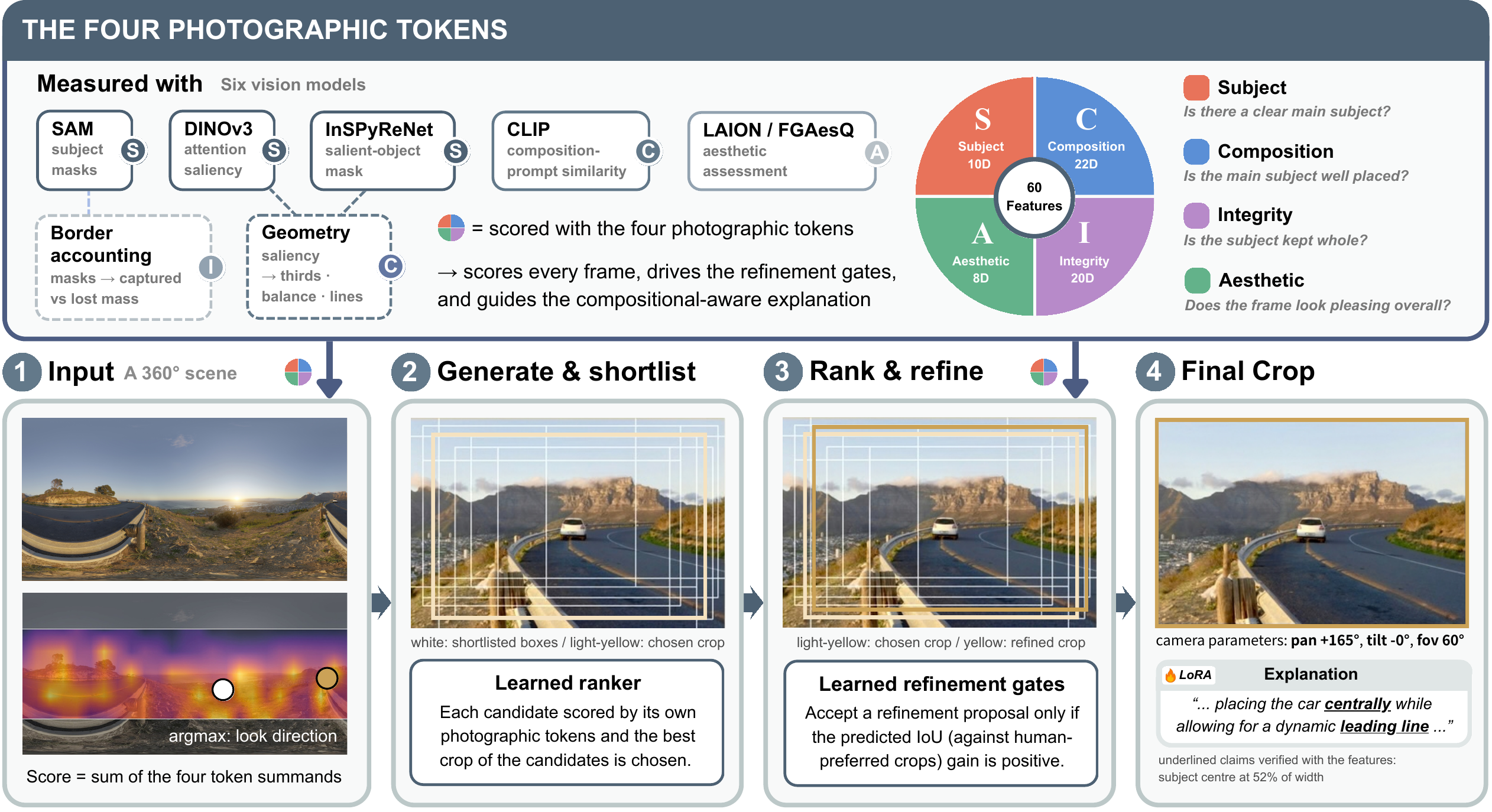}
    
    \vspace{-8pt}
    \caption{\textbf{One set of photographic tokens, three uses.} Each token yields one assessment score for a given frame, allowing the decision layer's output to be separated into four interpretable metrics. These drive our entire pipeline by (1) scoring all view directions in a 360\textdegree{}~scene, (2) ranking shortlisted framings and gating each refinement step, and (3) anchoring the fine-tuned MLLM's explanations. An optional generalist MLLM acts only as a consultant during refinement.}
    \label{fig:methodoverview}
\vspace{-1mm}
\end{figure}

\vspace{-2mm}
\section{Related Work}

\vspace{-2mm}
\paragraph{Image Cropping.}
Early croppers optimize hand-designed saliency and composition energies~\citep{nishiyama2009sensation,liu2010optimizing,chen2016automatic}. Generate-and-rank methods score candidate boxes using expert edits, professional photographs, saliency partitioning, or reinforcement learning~\citep{yan2013change,chen2017compose,wei2018goodview,lu2019listwise,shin2024order,wang2017deepcropping,wang2019deepnetwork,li2018a2rl,li2019a3rl,hamara2025efficient}. End-to-end methods directly regress crops using cascaded regression, grid anchors, score maps, meta-learning, or set prediction~\citep{guo2018automatic,zeng2019grid,zeng2022gaicd,tu2020asm,li2020mars,jia2022rethinking,wang2023spatial,pan2024pseudo,pan2023c2c}, often incorporating relational, boundary-aware, subject-aware, or synthesized composition signals~\citep{li2020cgs,hong2021cacnet,pan2021transview,liu2023unic,zhu2025attribute,wong2025aescrop,zhang2022humancentric,yang2023sacnet,hong2024gencrop,su2024s2cnet,gu2025videoreframe}. Their fixed objectives make the scoring evidence difficult to inspect. MLLM-based croppers use vision-language priors and fine-tuning or prompting for aesthetic reasoning, increasingly producing rationales, composition categories, or capture-time guidance alongside boxes~\citep{zhong2023clipcrop,yuan2024vlpcrop,lee2025cropper,du2026venus,zhang2026procrop,jiang2025multimodal,yuan2026humancrop,yang2026comex,viriyavisuthisakul2025xai,dong2026crop,you2026photoframer,li2026shuttermuse}. Photographic reasoning remains challenging for MLLMs~\citep{fang2025camerabench}. PhotoGuider generates candidates from self-supervised attention~\citep{caron2021emerging,simeoni2026dinov3} and ranks them through four photographic tokens with explicitly learned weights. Its MLLM consultant makes bounded proposals, aggregated following the principle of self-consistency~\citep{wang2022self}.

\vspace{-2mm}
\paragraph{Aesthetic Assessment and Explainable Critique.}
Aesthetic scoring has progressed from regressors trained on crowd-rated photographs~\citep{murray2012ava,talebi2018nima,he2022rethinking,huang2024multi} to CLIP- and LMM-based models~\citep{wang2023exploring,ke2023vila,wu2024qalign}. Composition datasets annotate thirds, symmetry, and triangles~\citep{lee2018photographic,zhang2021image,zhao2025composition}, providing reasoning priors for VLM croppers~\citep{dong2026crop}. Early critique generation used photographers' comments~\citep{chang2017aesthetic}. MLLMs generate aesthetic critiques at scale~\citep{huang2024aesbench,huang2024aesexpert,zhou2024uniaa,qi2025photographerseye}, assess quality through reinforcement learning~\citep{li2026q}, or use interpretable concept subspaces~\citep{liu2026concepts}. Their free-form rationales receive limited checks against measurable image properties, raising concerns about hallucination and faithfulness~\citep{li2023evaluating,guan2024hallusionbench,jacovi2020towards}. PhotoGuider uses aesthetic predictions and composition measurements as token inputs and filters explainer training drafts against these measurements, following a rationale-distillation approach akin to rejection sampling~\citep{zelikman2022star,hsieh2023distilling}.

\vspace{-2mm}
\paragraph{View Selection in 360{\upshape\textdegree} Scenes.}
Prior work extracts normal-field-of-view trajectories from panoramic video using learned videography, piloting agents, hyperlapse, highlight ranking, summarization, or saliency-guided camera paths~\citep{su2016pano2vid,su2017making,hu2017deep,lai2017semantic,yu2018deep,lee2018memory,kang2019interactive,wang2020transitioning360,chen2024saliency}. Other work optimizes projection, snap angles, or capture guidance~\citep{kim2017automatic,xiong2018snapangle,rao2022shoot360,su2021viewadjust,li2025spas}, predicts 360{\upshape\textdegree} viewer attention~\citep{rai2017dataset,sitzmann2018saliency,zhang2018saliency}, or steers real and reconstructed cameras with aesthetic rewards~\citep{fang2017creatism,xiong2017drone,alzayer2021autophoto,limoyo2024photobot,uchida20253dview,yao2025ppc,tang2026aestheticfield,jiang2026shutter,xu2026aes3d}. Saliency alone is insufficient for aesthetic view selection~\citep{chen2024saliency}, and existing view scores lack interpretability or conflate framing with camera-pose objectives. PhotoGuider uses the same 60 measurements in a LightGBM framing ranker and an additive direction scorer. It selects a direction, reprojects the view, and refines the crop, with shared measurements supporting view selection, ranking, and explanation.

\vspace{-2mm}
\paragraph{Interpretable Decision Layers.}
Generalized additive models provide per-feature contributions by construction, including boosted GA$^2$Ms, explainable boosting machines~\citep{lou2013accurate,nori2019interpretml}, and neural variants~\citep{agarwal2021neural,chang2021node,radenovic2022neural}. Gradient-boosted trees perform well on tabular data~\citep{ke2017lightgbm,grinsztajn2022trees}, support LambdaRank objectives~\citep{burges2010ranknet}, and admit isotonic calibration~\citep{zadrozny2002transforming,niculescu2005predicting}. PhotoGuider uses a token-grouped neural additive view scorer with pairwise terms and a calibrated LightGBM ranker and gates. TreeSHAP~\citep{lundberg2020treeshap} provides ranker attributions, with a secant approximation for the calibrated head (Appendix~\ref{a1}).

\vspace{-2mm}
\section{Method}
\label{method}

\vspace{-2mm}
\subsection{Overview}\label{method:3-1}
\vspace{-2mm}

\begin{figure}[!t]
    \centering
\vspace{-1mm}
    \includegraphics[width=1.00\linewidth]{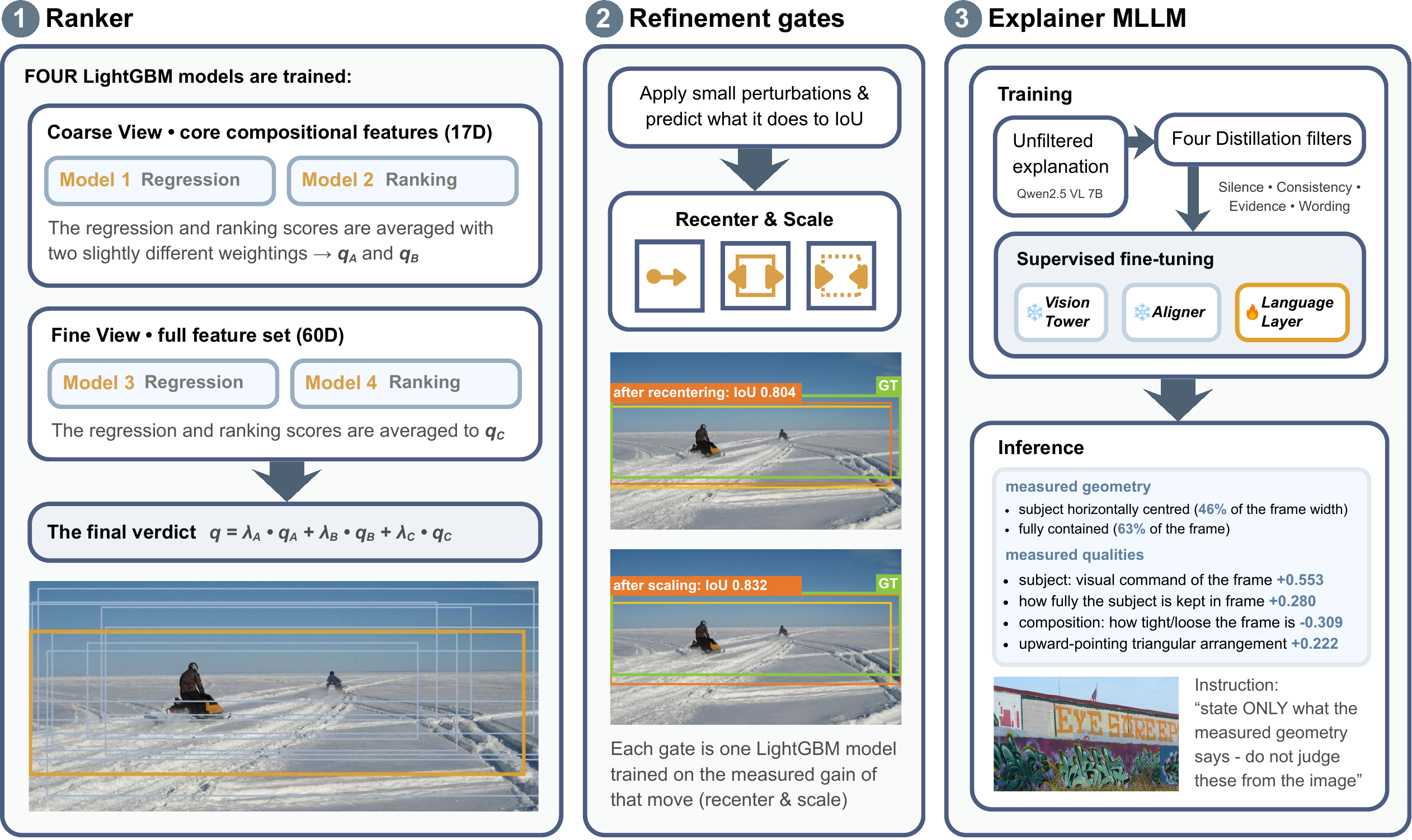}
    
    \vspace{-8pt}
    \caption{\textbf{The three learned components of PhotoGuider.} We use single ground-truth  crops from FCDB and top-rated crops from GAICD, with labels defined by the overlap (IoU) between candidate framings and reference crops. \textbf{(1) Ranker.} Candidates are scored using four LightGBM models (two regression and two ranking) operating on two feature tiers: a 17D core set (saliency, geometry) and a 60D full set (adding SAM object semantics, aesthetic scores, and CLIP prompts). Blending these model outputs yields three quality scores ($q_A, q_B, q_C$) to determine the optimal crop. \textbf{(2) Refinement Gates.} Six LightGBM gates adjust the selected crop by predicting IoU gains: one gate recenters the frame (via rule-of-thirds and CLIP-attention), four evaluate expansion (based on subject coverage, composition, scale-aesthetic variance, and golden ratio alignment), and a final gate shrinks the frame if no expansion triggers. \textbf{(3) Explainer MLLM.} It generates explanations conditioned on the crop, measured geometry, and ranker attributions. Training data drafts are filtered via four strict quality checks (unmasked geometry, numerical consistency, feature citations, and vocabulary validity).
}
    \label{fig:trainingoverview2}
\vspace{-1mm}
\end{figure}

A photographer arriving at a scene acts in four steps: \textit{decide which way to point the camera},
\textit{compose the shot}, \textit{polish it}, and, when possible, \textit{ask a second pair of eyes}. PhotoGuider performs these steps as follows: Stage (a) extends the 2D cropper into a complete 360{\upshape\textdegree}~framing system by scoring candidate look directions across the panorama using shared assessments and reprojecting the optimal direction into a working view (Sec.~\ref{method:3-7}). Stage (b) generates, filters, and ranks candidate framings to select a top proposal (Secs.~\ref{method:3-2} and~\ref{method:3-3}). Stage (c) applies spatial adjustments when predicted beneficial by a learned gate (Sec.~\ref{method:3-4}). Together, stages (b) and (c) constitute the \textbf{fast mode}, and the \textbf{full mode} further incorporates stage (d), which consults a specialized MLLM within bounded constraints (Sec.~\ref{method:3-5}). Finally, composition-aware explanations are derived from the same assessments (Sec.~\ref{method:3-6}). The method and training overviews are shown in Fig.~\ref{fig:methodoverview} and Fig.~\ref{fig:trainingoverview2}, respectively.

\begin{figure}[!t]
    \centering
\vspace{-1mm}
    \includegraphics[width=1.00\linewidth]{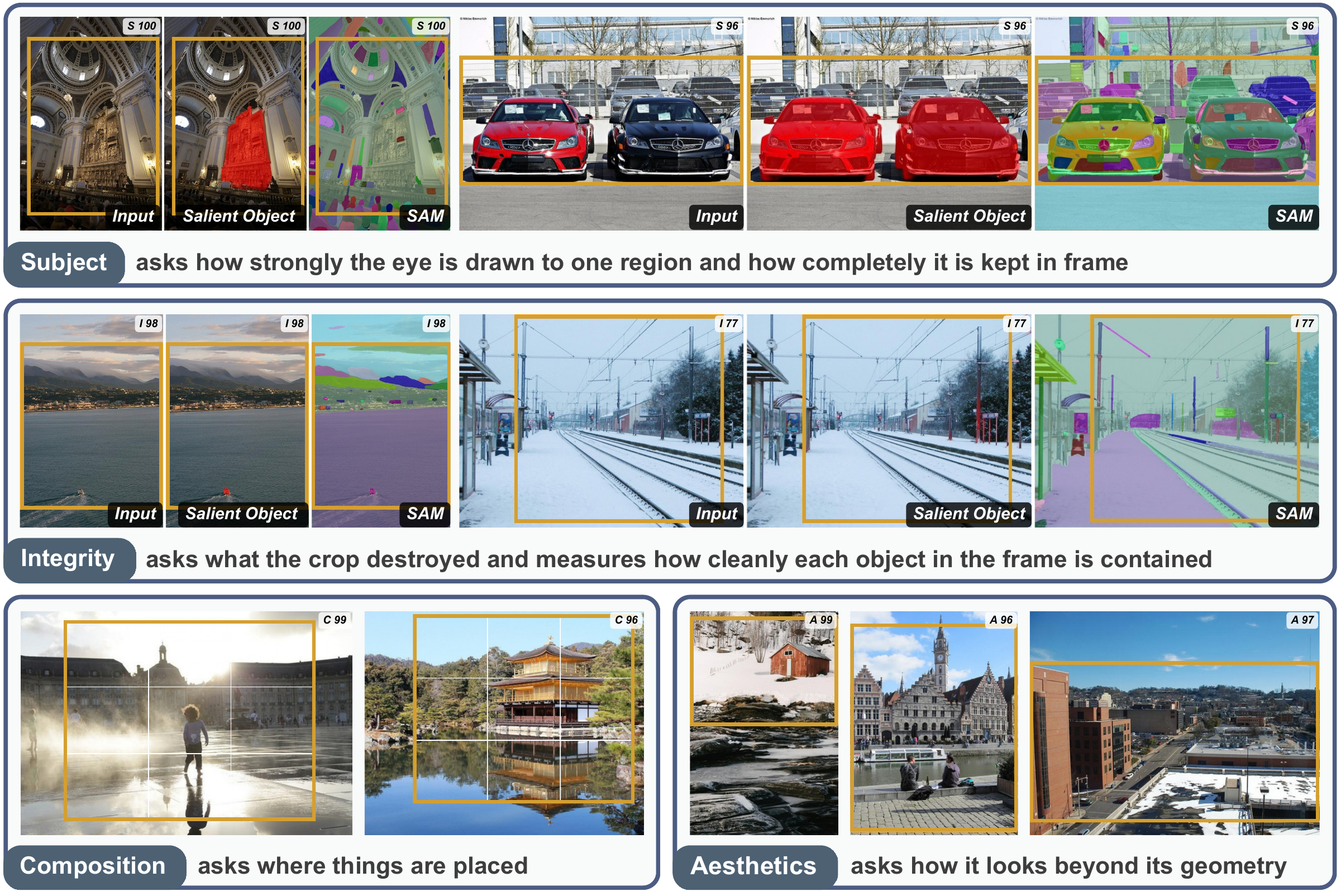}
    
    \vspace{-8pt}
    \caption{\textbf{Feature Partitioning and Token Definitions.} The four tokens partition the 60 extracted features into disjoint subsets. Each reported score represents one such component, expressed as a percentile against the same quantity derived from training crops that match the human annotations most closely. The assessed frame is highlighted by the yellow box.}
    \label{fig:tokenoverview}
\vspace{-3mm}
\end{figure}

\vspace{-2mm}
\paragraph{The four photographic tokens}

Each token is extracted by the model whose mechanism matches it. Subject is answered by three sources: the self-supervised attention map of DINOv3~\citep{simeoni2026dinov3} highlights regions that naturally draw visual focus, InSPyReNet \citep{kim2022inspyrenet} extracts the salient-object mask, and SAM \citep{kirillov2023sam} segments the distinct objects and semantic regions comprising the frame. Composition is evaluated by CLIP \citep{radford2021clip}, measuring the alignment between the crop and compositional geometric attributes like the rule of thirds and visual lines computed over the saliency map. Integrity is quantified via geometric and boundary statistics relating the segmented objects and subject mask to the crop border. Aesthetics is predicted using two learned aesthetic assessment models, LAION \citep{schuhmann2022laion5b} and FGAesQ \citep{yang2026fine}. Figure~\ref{fig:tokenoverview} illustrates the four tokens. Each description below states what a token measures; the definitions of every feature are provided in Appendix~\ref{a1}.

\vspace{-1mm}
\textbf{Subject (10D).} Evaluates focal attention and object completeness using three models: DINOv3 attention captures focal strength (3D), InSPyReNet saliency quantifies subject coverage and prominence (2D), and SAM provides frame coverage, spatial placement, and cross-model agreement with InSPyReNet (4D), supplemented by subject scale (1D).

\vspace{-1mm}
\textbf{Integrity (20D).} Measures information preservation and structural disruption using SAM-derived representation. Encoder features (9D) quantify visual information density, object signal distribution, crop-versus-background contrast, whole-image baseline, and shot scale. Decoder features (11D) capture object-level composition, including object count, dominant area shares, size disparity, boundary containment cleanliness, edge margins, and depth layering.

\vspace{-1mm}
\textbf{Composition (22D).} Spatial layout is parameterized across three sources: saliency geometry (10D) covering rule-of-thirds, symmetry, triangles, balance, offset, tightness, and aspect ratio; SAM object parsing (3D) evaluating grid alignment and center-versus-thirds bias; and CLIP features (9D) measuring eight canonical cues plus peak match score.

\vspace{-1mm}
\textbf{Aesthetics (8D).} Captures visual appeal via two learned taste models (2D LAION crowd appeal scores and 1D FGAesQ expert judgment), 1D overall saturation, and 4D SAM-derived subject-to-background contrast metrics in saturation, brightness, Lab color, and subject color richness.

\vspace{-2mm}
\subsection{Candidate generation and shortlisting}\label{method:3-2}
\vspace{-2mm}

Candidate generation only uses the visual signals from the DINOv3 attention map, the InSPyReNet salient-object mask, and box geometry. Given a target aspect ratio ${asp}^{*}$, eight complementary sources generate a candidate pool $\mathcal{P}$ of approximately 300 boxes, comprising a re-aspected base geometric grid, subject mask and top saliency anchors paired with seven medium scales, a multi-scale refinement pass over top-covering candidates, and four composition-guided generators (diagonal, negative space, leading line, and boundary). In unconstrained settings without a fixed ratio, a sampling grid of seven scales and 36 positions spans eight aspect ratios, including seven standard ratios ($16{:}9$, $3{:}2$, $4{:}3$, $1{:}1$, $3{:}4$, $2{:}3$, and $9{:}16$) alongside the native image ratio. A determinantal point process (DPP)~\citep{kulesza2012dpp,kulesza2011kdpp,chen2018fastgreedy} selects ten framings ($K = 10$) that are individually strong and mutually distinct. Candidate quality is defined as:
\begingroup
\small
\begin{equation}
q_{\mathrm{dpp}}(c) = \bigl(q_{\mathrm{LGBM}}(c) + \lambda_{v} s_{\mathrm{sal}}(c)\bigr) \cdot \max\Bigl(0, 1 - \lambda_{a}\Bigl|\log\frac{\mathrm{asp}(c)}{\mathrm{asp}^{*}}\Bigr|\Bigr),
\end{equation}
\endgroup
where $q_{\mathrm{LGBM}}$ is a LightGBM~\citep{ke2017lightgbm,grinsztajn2022trees,mcelfresh2023neuralvstrees,holzmuller2024betterbydefault,gorishniy2024tabr} quality regressor trained on FCDB over saliency and geometry features, $s_{\mathrm{sal}}$ denotes peak DINOv3 saliency within the crop, and the final term penalizes aspect ratio deviations from the target $\mathrm{asp}^{*}$ (evaluating to one in free-form mode). The diversity kernel is $\exp(-\beta(1-\mathrm{IoU}))$ with a same-source penalty. Greedy MAP approximation~\citep{chen2018fastgreedy} extracts the shortlist $\mathcal{K}$. The top three candidates $\mathcal{K}_3$ are forwarded to the main ranker (Sec.~\ref{method:3-3}), while all ten candidates provide normalization context.

\vspace{-2mm}
\subsection{Ranking}\label{method:3-3}
\vspace{-2mm}

The weighting of the four questions varies by photograph. A LightGBM ranker learns it on the human-favored framings, supervised by IoU, and returns the chosen frame:
\begingroup
\small
\begin{equation}
q(c)=\lambda_{A}\,q_{A}(c)+\lambda_{B}\,q_{B}(c)+\lambda_{C}\,q_{C}(c),\qquad
c^{*}=\arg\max_{c\in\mathcal{K}_3} q(c),
\end{equation}
\endgroup
where $\lambda_{A}+\lambda_{B}+\lambda_{C}=1$ are mixing weights. Two regression-and-ranking model pairs use different feature sets. The first pair uses the \textbf{17-dimensional core compositional set} (the saliency measures of Subject, the geometric rules of Composition, and one color term). Its regression and ranking outputs are combined with two sets of weights to obtain $q_{A}$ and $q_{B}$. The second pair uses the \textbf{full 60-dimensional set}, applies isotonic calibration to its regression head, and produces $q_{C}$. Each component is min-max
normalized within the image. Pool-relative features are computed over all ten boxes and then
frozen; normalization and the $\arg\max$ run on $\mathcal{K}_3$. Token attributions are exact for three uncalibrated heads carrying a combined weight of 0.88; the calibrated head (weight 0.12) uses a secant approximation (Appendix~\ref{a1}).

\vspace{-2mm}
\subsection{Gated Refinement}\label{method:3-4}
\vspace{-2mm}

Three sequential spatial moves adjust the chosen frame $c^{*}$ into the refined frame $b$. Each move is gated by a LightGBM regressor predicting the expected IoU gain on the current image, firing only when the predicted gain exceeds a dedicated threshold. In the recenter stage, two center targets compete for alignment: $t_{\mathrm{thirds}}$, representing the saliency centroid aligned with the nearest rule-of-thirds point, and $t_{\mathrm{cam}}$, representing the centroid of CLIP compositional attention. The gate predicts the IoU advantage $\hat{m}$ of the CLIP target, computes a weight $w = \mathrm{clip}(\hat{m} / T_{g}, 0, 1)$, and shifts the frame center by a fraction $\beta$ toward $t = (1 - w) t_{\mathrm{thirds}} + w t_{\mathrm{cam}}$. In the Expand stage, four independent gates evaluate the gain $\hat{g}_{k}(s)$ of enlarging the frame by scale factor $s \in \mathcal{S}$ centered at a fixed ratio. Each gate assesses a distinct signal: subject coverage, CLIP wide-versus-tight response, aesthetic score variation under enlargement, and subject proximity to the nearest golden-section point. A gate fires when $\max_{s} \hat{g}_{k}(s) > \tau_{e}$, and the frame adopts the scale factor providing the highest predicted gain among all fired gates. In the shrink stage, if no expansion gate fires, a fallback gate evaluates shot-scale similarity via the cosine similarity between SAM embeddings of the frame and full image. Above threshold $\tau_{s}$, the gate tightens frames using scale factor $s \in \mathcal{S}_{\mathrm{shr}}$.

\vspace{-2mm}
\subsection{MLLM as Consultant}\label{method:3-5}
\vspace{-2mm}

Venus~\citep{du2026venus} acts as an MLLM consultant to propose candidate boxes without scoring or making decisions in the pipeline. A primary prompt undergoes $1 + m$ decodes (one greedy decoding alongside $m$ temperature-sampled outputs), where the coordinate-wise median establishes the core opinion $\tilde{v}_{\mathrm{c}}$. Combining outputs from $M$ paraphrased prompts extends this to a broader opinion $\tilde{v}_{\mathrm{f}}$ via median aggregation. Each candidate $c_i$ aligning with the consultant ($\mathrm{IoU}(c_i, \tilde{v}_{\mathrm{c}}) \ge 0.5$) receives a twin candidate adjusted toward $\tilde{v}_{\mathrm{c}}$ within fixed bounds. After recomputing all 60 features, the ranker compares the original and twin candidates, switching only given a significant score margin. If the refined frame $b$ agrees with the broader opinion, its center and area are interpolated.

\vspace{-2mm}
\subsection{Explanation}\label{method:3-6}
\vspace{-2mm}

Our explainer cites the quantitative features used by the ranker. Token assessments and feature attributions originate from TreeSHAP~\citep{lundberg2017shap,lundberg2020treeshap} decomposition of the ranker's output, while geometric statements are computed from the same vision features. Qwen2.5-VL-7B~\citep{bai2025qwen25vl} is the teacher model used to draft explanations for all training images. Four distillation filters admit valid drafts: \emph{Silence} (rejecting geometric claims lacking valid subject masks), \emph{Consistency} (rejecting values conflicting with physical measurements), \emph{Evidence} (enforcing feature citations beyond basic geometry), and \emph{Wording} (eliminating banned phrasing or incorrect languages). The student model learns exclusively from admitted drafts via QLoRA~\citep{dettmers2023qlora,hu2022lora}.

\vspace{-2mm}
\subsection{Generalization to 360{\upshape\textdegree}~scenes}\label{method:3-7}

\vspace{-1mm}
\textbf{Coarse: choosing a direction.} Perspective views are reprojected on an overlapping
grid of pan and tilt angles, and the same 60 features are extracted for each view
exactly as for a crop. An {additive view scorer} maps the features of a view to a
composition score. It is trained on the GAICD \citep{zeng2022gaicd} training set, whose mean opinion scores rate candidate frames. The scoring model is a token-grouped neural additive network whose output is the sum of four token-specific summands, attributing the strength of each candidate direction to its constituent tokens. The direction with the highest aggregate score is selected as the optimal view.

\vspace{-1mm}
\textbf{Fine: framing in the chosen direction.} The selected direction is reprojected into
a wider working view, which then enters the cropping pipeline described in Secs.~\ref{method:3-2}, \ref{method:3-3}, \ref{method:3-4}, and \ref{method:3-5}. The resulting box is converted through the working view's pinhole model into camera
parameters, pan, tilt, field of view and aspect ratio, and the photograph is re-rendered from the panorama at full resolution.

\vspace{-2mm}
\section{Experiments}
\label{experiment}

\definecolor{lightred}{RGB}{253,191,191}
\definecolor{lightorange}{RGB}{255,223,191}
\definecolor{lightyellow}{RGB}{254,240,198}

\begin{table*}[t]
\vspace{-6mm}
\small
\centering
\caption{\textbf{Comparison with state-of-the-art methods on FCDB and FLMS datasets.}}
\label{tab:sota_comparison}
\renewcommand{\arraystretch}{1.1}

\vspace{5pt}

\begin{tabular}{l | c | cccccc}
\toprule
\multirow{2}{*}{Method} &
\multirow{2}{*}{Publication} &
\multicolumn{2}{c}{FCDB} &
\multicolumn{2}{c}{FLMS} &
\multicolumn{2}{c}{Average} \\
\cmidrule(lr){3-4}
\cmidrule(lr){5-6}
\cmidrule(lr){7-8}

& &
BDE$\downarrow$ & IoU$\uparrow$ &
BDE$\downarrow$ & IoU$\uparrow$ &
BDE$\downarrow$ & IoU$\uparrow$ \\
\midrule

DIC \cite{wang2017deepcropping} & ICCV 2017 & 0.090 & 0.630 & 0.057 & 0.810 & 0.074 & 0.720 \\

DIC$^{*}$ \citep{wang2019deepnetwork} & TPAMI 2019 & 0.080 & 0.650 & 0.052 & 0.830 & 0.066 & 0.740 \\

A2RL \citep{li2018a2rl} & CVPR 2018 & 0.089 & 0.664 & 0.045 & 0.821 & 0.067 & 0.743 \\

A3RL \citep{li2019a3rl} & TIP 2019 & 0.077 & 0.696 & --    & 0.839 &
-- &
0.768 \\

VPN \citep{wei2018goodview} & CVPR 2018 & {0.073} & {0.711} & 0.044 & 0.836 &
0.059 &
0.774 \\

CACNet \citep{hong2021cacnet} & CVPR 2021 &
0.074 &
0.702 &
\cellcolor{lightyellow}{0.037} &
{0.841} &
0.056 &
0.772 \\

GAIC \citep{zeng2022gaicd} & TPAMI 2022 & 0.081 & 0.674 & 0.041 & 0.834 & 
0.061 & 
0.754 \\

CAGR \citep{zhu2025attribute} & TOMM 2025 &
{0.067} &
{0.721} &
\cellcolor{lightorange}{0.033} &
\cellcolor{lightyellow}{0.851} & 
\cellcolor{lightyellow}0.050 & 
0.786 \\

InstructCrop \citep{sheng2025instructcrop} & MM 2025 &
{0.070} &
{0.716} &
\cellcolor{lightorange}{0.033} &
{0.847} & 
0.052 & 
0.782 \\

Cropper \citep{lee2025cropper} & CVPR 2025 & 
\cellcolor{lightyellow}0.062 & 
\cellcolor{lightyellow}0.756 & 
0.041 & 
0.821 & 
0.052 &
0.789 \\

Venus \citep{du2026venus} & CVPR 2026 & 
0.071 & 
0.713 & 
\cellcolor{lightred}0.030 & 
\cellcolor{lightred}0.870 & 
0.051 &
\cellcolor{lightyellow}0.792 \\ 

CPIC \citep{zhang2026discrete} & arXiv 2026 & 
\cellcolor{lightorange}0.058 & 
0.751 & 
\cellcolor{lightyellow}0.037 & 
0.842 & 
\cellcolor{lightorange}0.048 &
\cellcolor{lightorange}0.797 \\

\midrule

PhotoGuider (fast)  & 2026 & 
\cellcolor{lightorange}0.058 & 
\cellcolor{lightorange}0.759 & 
0.041 & 
0.835 &
\cellcolor{lightyellow}0.050 &
\cellcolor{lightorange}0.797 \\

PhotoGuider (full) \ \ \ & 2026 & 
\cellcolor{lightred}0.056 & 
\cellcolor{lightred}0.766 & 
\cellcolor{lightorange}0.033 & 
\cellcolor{lightorange}0.856 &
\cellcolor{lightred}0.044 &
\cellcolor{lightred}0.811 \\

\bottomrule
\end{tabular}
\vspace{\dimexpr 2mm-8pt\relax}
\end{table*}

\vspace{-2mm}
\subsection{Datasets and Metrics}\label{sec:dataset}
\vspace{-2mm}

Following the protocol of Cropper \citep{lee2025cropper} and Venus \citep{du2026venus}, we evaluate PhotoGuider across two cropping benchmarks: the FCDB testing set (348 images) for aspect ratio-aware cropping, and FLMS (500 images) for free-form cropping. Training samples are collected from the FCDB training set with single ground-truth crops (1395 images) and GAICD with top-rated crops (1000 images). We report the mean intersection-over-union (mIoU) and the boundary displacement error (BDE). Implementation details are provided in Appendix~\ref{a5}.

\vspace{-2mm}
\subsection{Comparison with Baselines}\label{sec:mainresults}
\vspace{-2mm}

\begin{figure}[!t]
\vspace{-1mm}
    \centering
    \includegraphics[width=1.00\linewidth]{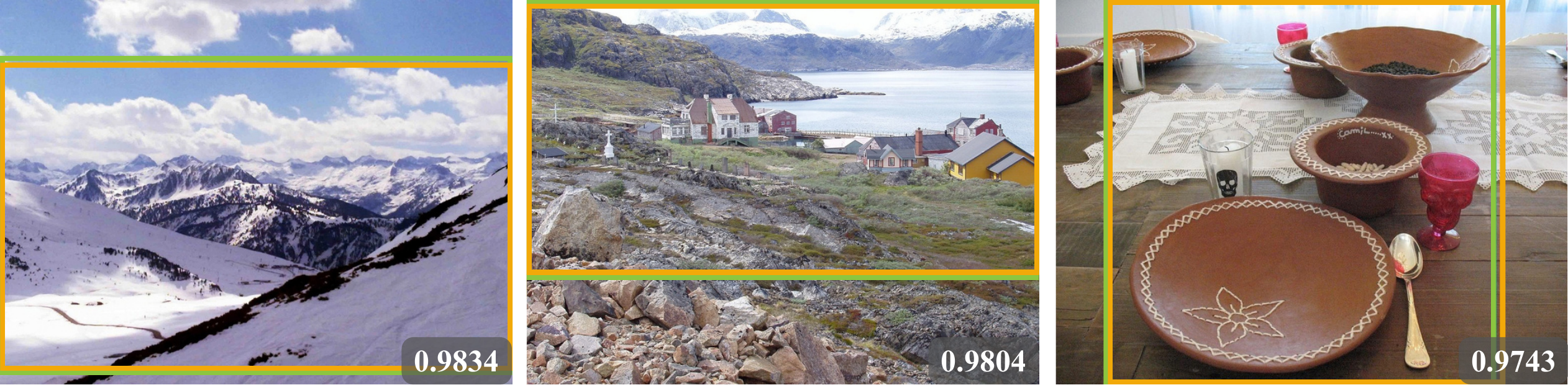}
    
    \vspace{-8pt}
    \caption{\textbf{Examples from PhotoGuider on FCDB~\citep{chen2017quantitative} and FLMS~\citep{fang2014automatic} testing images.} Green boxes are ground-truth crops from both datasets, and yellow boxes show crops from PhotoGuider. Bottom-right: IoU between the predicted and ground-truth crops.} 
    \label{fig:croppingvisualization}
\vspace{-1mm}
\end{figure}

We compare the 2D cropping performance of PhotoGuider on the FCDB and FLMS datasets against state-of-the-art baselines, including DIC \citep{wang2017deepcropping}, DIC$^{*}$ \citep{wang2019deepnetwork}, A2RL \citep{li2018a2rl}, A3RL \citep{li2019a3rl}, VPN \citep{wei2018goodview}, CACNet \citep{hong2021cacnet}, GAIC \citep{zeng2022gaicd}, CAGR \citep{zhu2025attribute}, InstructCrop \citep{sheng2025instructcrop}, Cropper \citep{lee2025cropper}, and Venus \citep{du2026venus}. The results are summarized in Table~\ref{tab:sota_comparison}. We evaluate FCDB and FLMS using the same trained checkpoint, without dataset-specific retraining. PhotoGuider achieves the highest mean mIoU and lowest mean BDE averaged across the two benchmarks. The example crops from PhotoGuider are shown in Fig.~\ref{fig:croppingvisualization}. For each drop-one and keep-only configuration, we retrain the ranker using the corresponding feature subset and evaluate fast mode. The results are shown in Table~\ref{tab:token_ablation}. Subject and Composition provide the strongest individual feature groups, while combining all four tokens achieves the best overall cropping performance. Integrity and Aesthetics receive lower weights in the ranker because candidate shortlisting drastically reduces their variance within candidate pools. Since candidates are anchored on the salient subject, subject capture exceeds $0.95$ for $64\%$ of FCDB and all FLMS shortlist frames, varying within a pool at only $0.4$ of its global spread (compared to $0.9$ for rule-of-thirds alignment). Instead, these features operate in downstream modules: the coverage and shrink gates use integrity measures, while the aesthetic delta gate evaluates aesthetic score variations. The integrity and aesthetic assessments are the statements the explainer follows most faithfully. To also validate the performance of our fine-tuned explainer, a rule-based detector checks every numeric and directional claim in an explanation against the injected measurement and flags a contradiction when they
disagree. With the same prompt and detector on the testing images, fine-tuning lowered the detector-measured contradiction rate of Qwen2.5-VL-7B from 6.31\% to 1.55\% (Appendix~\ref{a4}). Moreover, we conducted a double-blind, three-way user study. We evaluated PhotoGuider against Venus~\cite{du2026venus} and ArtiMuse~\cite{cao2026artimuse} across five test images (two from FCDB, three from FLMS) using identical crops. 50 participants viewed the three candidate explanations with model names removed and order randomized, selecting the text that best explained the crop's composition. PhotoGuider’s explanations were preferred most often, capturing 40.0\% of the 250 total votes (ahead of Venus at 32.8\% and ArtiMuse at 27.2\%). This confirms that users judging image composition favor explanations grounded in quantitative compositional metrics at least as strongly as those produced by a specialized aesthetic MLLM. Details can be found in Appendix~\ref{a7}.

\vspace{-2mm}
\begin{table}[t]
\centering
\small
\renewcommand{\arraystretch}{1.1}
\setlength{\abovecaptionskip}{2pt}
\caption{\textbf{Photographic Token ablation of PhotoGuider in fast mode.}}

\vspace{5pt}

\begin{tabular}{l | cc cc | cc cc}
\toprule
& \multicolumn{4}{c|}{Drop-one} & \multicolumn{4}{c}{Keep-only} \\
\cmidrule(lr){2-5}\cmidrule(lr){6-9}
& \multicolumn{2}{c}{FCDB} & \multicolumn{2}{c|}{FLMS} & \multicolumn{2}{c}{FCDB} & \multicolumn{2}{c}{FLMS} \\
Token & BDE$\downarrow$ & IoU$\uparrow$ & BDE$\downarrow$ & IoU$\uparrow$ & BDE$\downarrow$ & IoU$\uparrow$ & BDE$\downarrow$ & IoU$\uparrow$ \\
\midrule
Subject (S)      & {0.063} & {0.740} & \textbf{0.046} & \textbf{0.821} & \textbf{0.064} & \textbf{0.735} & \textbf{0.042} & \textbf{0.834} \\
Composition (C) \ & \textbf{0.064} & \textbf{0.737} & {0.042} & {0.834} & \textbf{0.064} & \textbf{0.735} & 0.046 & 0.820 \\
Integrity (I)    & {0.060} & {0.754} & {0.042} & {0.834} & 0.068 & 0.718 & 0.044 & 0.826 \\
Aesthetics (A)  & {0.060} & 0.753 & {0.042} & {0.834} & 0.070 & 0.714 & 0.052 & 0.792 \\

\bottomrule
\end{tabular}
\label{tab:token_ablation}
\end{table}

\subsection{Performance of 360{\upshape\textdegree}~Framing}\label{sec:panoramic-framing}
\vspace{-2mm}

\begin{figure}[!t]
    \centering
    \includegraphics[width=1.00\linewidth]{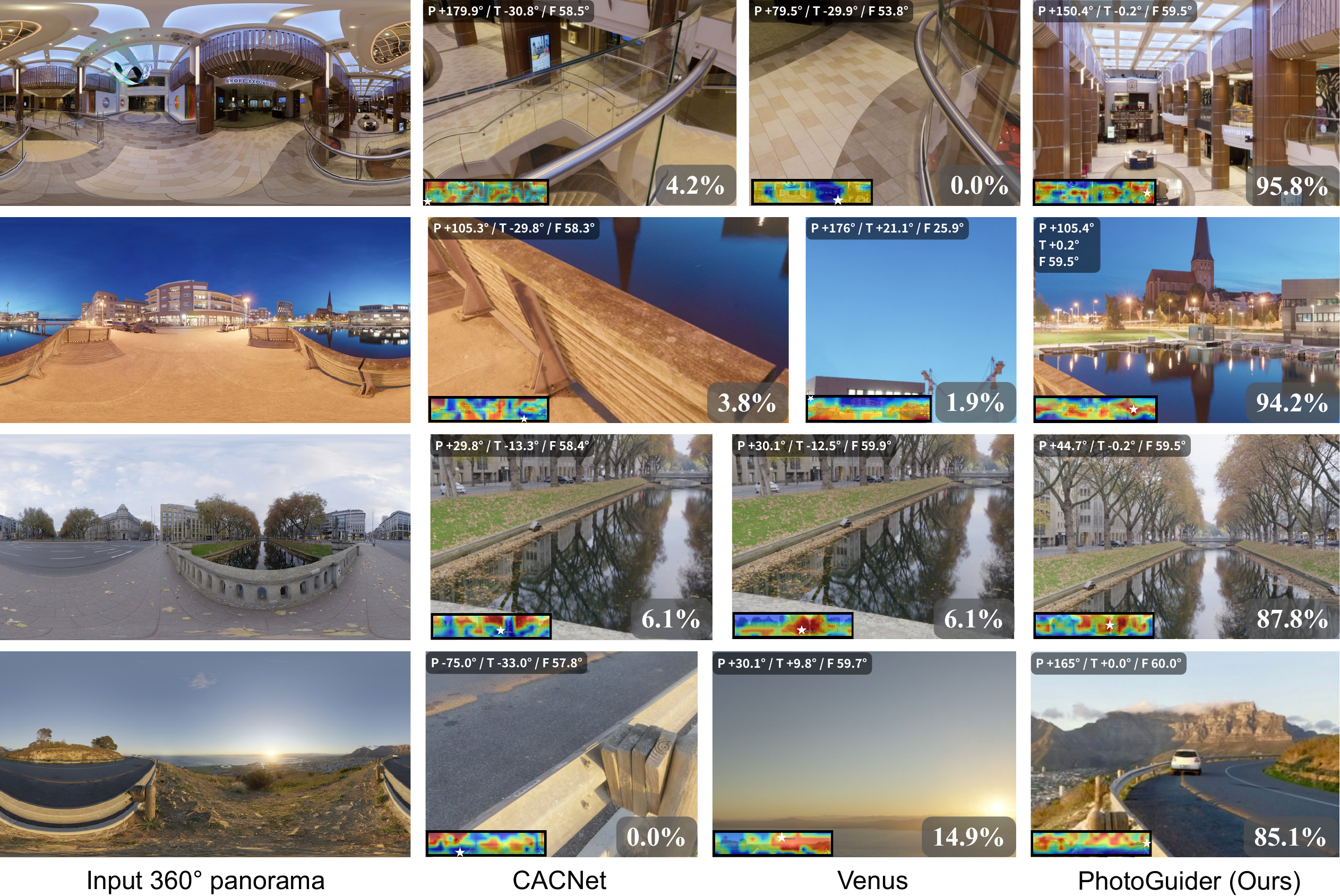}
    
    \vspace{-8pt}
    \caption{\textbf{Visual comparison of 360\textdegree{}~view framing from the  Poly Haven HDRI library.} Score heatmaps are produced by each method across 120 look directions (24 pan × 5 tilt). Scores are displayed as percentile ranks, with red denoting highly favored directions and a star marking the peak score, alongside the corresponding rendered photograph. Each panel labels the chosen camera parameters (pan / tilt / field of view) and shows the percentage of participants who rated it first. Across 100 scenes, the first-place preference shares are 19.5\%, 27.3\%, and 53.2\%, respectively.} 
    \label{fig:panocropping}
    \vspace{-8pt} 
\end{figure}

\begin{wraptable}{r}{0.49\textwidth}\vspace{-1.5\baselineskip}\vspace{-3pt}
\small
\centering
\caption{\textbf{User study.} Rank-1/3 report the first- and third-place shares, and Scenes Won counts scenes receiving the most first-place votes. PhotoGuider gets 53.2\% first-place and 15.2\% third-place rankings, wins 59 of 100 scenes, and is preferred head-to-head over Venus in 65.2\% and CACNet in 72.8\% of comparisons.}

\vspace{5pt}

\label{tab:user-study}
\renewcommand{\arraystretch}{1.1}
\setlength{\tabcolsep}{1.5pt}
\begin{tabularx}{\linewidth}{@{}>{\raggedright\arraybackslash}p{0.22\linewidth} | *{4}{>{\centering\arraybackslash}X}@{}}
\toprule
\multirow{2}{*}{Method} & \multirow{2}{*}{Rank-1} & \multirow{2}{*}{Rank-3} & Mean & Scenes \\
& & & rank & won \\
\midrule
\textbf{Ours} & \textbf{53.2\%} & \textbf{15.2\%} & \textbf{1.62} & \textbf{59} \\
Venus & 27.3\% & 37.4\% & 2.10 & 26 \\
CACNet & 19.5\% & 47.4\% & 2.28 & 15 \\
\bottomrule
\end{tabularx}
\vspace{-10pt}
\end{wraptable}
We evaluate 360{\upshape\textdegree}~framing through a user study with 120 participants and 100 panoramic scenes. We collected 100 panoramic scenes from the Poly Haven library, covering diverse indoor and outdoor environments across both day and night conditions. We compare PhotoGuider (fast mode) against CACNet \citep{hong2021cacnet}, an end-to-end deep cropper, and Venus \citep{du2026venus}, an MLLM-based cropper. All methods follow an identical two-stage protocol using their respective scoring mechanisms. CACNet uses the maximum confidence score from its composition classifier, whereas Venus relies on its composition score (1--10 scale) decoded greedily with a fixed seed. The top-scoring view establishes the primary direction, after which each method crops its corresponding $60^\circ$ working view. The resulting bounding boxes are converted to camera parameters and re-rendered from the panorama. None of the three methods is trained on panoramas. Participants ranked the three anonymized framings for each scene in a randomized order. The participants provided 4,820 rankings. PhotoGuider received 53.2\% of first-place rankings, compared with 27.3\% for Venus and 19.5\% for CACNet (Table~\ref{tab:user-study}). Qualitative comparisons among the three methods are demonstrated in Fig.~\ref{fig:panocropping}.

\vspace{-2mm}
\section{Conclusion}
\label{conclusion}
\vspace{-2mm}

\textbf{PhotoGuider} models a photographer's framing process using four interpretable photographic tokens. These tokens drive a compact ranker for candidate selection, a gated refinement module, and an explainer providing grounded rationales. PhotoGuider achieves competitive performance across two benchmarks and, to our knowledge, is the first method to combine explicit composition-based direction selection in 360\textdegree{}~panoramas with refined framing, camera-parameter output, and measurement-conditioned explanations. User evaluations show clear preference for our crops over end-to-end and MLLM baselines. PhotoGuider combines view selection, crop refinement, and measurement-conditioned explanations in one pipeline.

\vspace{-1mm}
\textbf{Limitations.} Framing adjustments rely on discrete step fractions, approximating targets falling between sampled intervals. Also, view selection uses a single argmax, returning one frame even in scenes with multiple strong regions regardless of user preference. Future work will examine continuous refinement and user-guided view selection.

\bibliography{iclr2027_conference}

\begin{thebibliography}{139}
\providecommand{\natexlab}[1]{#1}
\providecommand{\url}[1]{\texttt{#1}}
\expandafter\ifx\csname urlstyle\endcsname\relax
  \providecommand{\doi}[1]{doi: #1}\else
  \providecommand{\doi}{doi: \begingroup \urlstyle{rm}\Url}\fi

\bibitem[Agarwal et~al.(2021)Agarwal, Melnick, Frosst, Zhang, Lengerich,
  Caruana, and Hinton]{agarwal2021neural}
Rishabh Agarwal, Levi Melnick, Nicholas Frosst, Xuezhou Zhang, Ben Lengerich,
  Rich Caruana, and Geoffrey~E. Hinton.
\newblock Neural additive models: Interpretable machine learning with neural
  nets.
\newblock \emph{Advances in neural information processing systems},
  34:\penalty0 4699--4711, 2021.

\bibitem[AlZayer et~al.(2021)AlZayer, Lin, and Bala]{alzayer2021autophoto}
Hadi AlZayer, Hubert Lin, and Kavita Bala.
\newblock Autophoto: Aesthetic photo capture using reinforcement learning.
\newblock In \emph{{IEEE/RSJ} International Conference on Intelligent Robots
  and Systems, {IROS} 2021, Prague, Czech Republic, September 27 - Oct. 1,
  2021}, pp.\  944--951. {IEEE}, 2021.
\newblock \doi{10.1109/IROS51168.2021.9636788}.

\bibitem[Bai et~al.(2025)Bai, Chen, Liu, Wang, Ge, Song, Dang, Wang, Wang,
  Tang, Zhong, Zhu, Yang, Li, Wan, Wang, Ding, Fu, Xu, Ye, Zhang, Xie, Cheng,
  Zhang, Yang, Xu, and Lin]{bai2025qwen25vl}
Shuai Bai, Keqin Chen, Xuejing Liu, Jialin Wang, Wenbin Ge, Sibo Song, Kai
  Dang, Peng Wang, Shijie Wang, Jun Tang, Humen Zhong, Yuanzhi Zhu, Mingkun
  Yang, Zhaohai Li, Jianqiang Wan, Pengfei Wang, Wei Ding, Zheren Fu, Yiheng
  Xu, Jiabo Ye, Xi~Zhang, Tianbao Xie, Zesen Cheng, Hang Zhang, Zhibo Yang,
  Haiyang Xu, and Junyang Lin.
\newblock Qwen2.5-vl technical report.
\newblock \emph{CoRR}, abs/2502.13923, 2025.
\newblock \doi{10.48550/ARXIV.2502.13923}.

\bibitem[Brooks et~al.(2019)Brooks, Mildenhall, Xue, Chen, Sharlet, and
  Barron]{brooks2019unprocessing}
Tim Brooks, Ben Mildenhall, Tianfan Xue, Jiawen Chen, Dillon Sharlet, and
  Jonathan~T. Barron.
\newblock Unprocessing images for learned raw denoising.
\newblock In \emph{Proceedings of the IEEE/CVF Conference on Computer Vision
  and Pattern Recognition (CVPR)}, pp.\  11036--11045, June 2019.

\bibitem[Burges(2010)]{burges2010ranknet}
Christopher J.~C. Burges.
\newblock From {RankNet} to {LambdaRank} to {LambdaMART}: An overview.
\newblock Technical Report MSR-TR-2010-82, Microsoft Research, 2010.

\bibitem[Cao et~al.(2026)Cao, Ma, Li, Li, Shao, Zhu, Zhou, Pu, Wu, Wang,
  et~al.]{cao2026artimuse}
Shuo Cao, Nan Ma, Jiayang Li, Xiaohui Li, Lihao Shao, Kaiwen Zhu, Yu~Zhou,
  Yuandong Pu, Jiarui Wu, Jiaquan Wang, et~al.
\newblock Artimuse: Fine-grained image aesthetics assessment with joint scoring
  and expert-level understanding.
\newblock In \emph{Proceedings of the IEEE/CVF Conference on Computer Vision
  and Pattern Recognition}, pp.\  15313--15322, 2026.

\bibitem[Caron et~al.(2021)Caron, Touvron, Misra, J{\'e}gou, Mairal,
  Bojanowski, and Joulin]{caron2021emerging}
Mathilde Caron, Hugo Touvron, Ishan Misra, Herv{\'e} J{\'e}gou, Julien Mairal,
  Piotr Bojanowski, and Armand Joulin.
\newblock Emerging properties in self-supervised vision transformers.
\newblock In \emph{2021 IEEE/CVF international conference on computer vision
  (ICCV)}, pp.\  9630--9640. IEEE, 2021.

\bibitem[Chang et~al.(2021)Chang, Caruana, and Goldenberg]{chang2021node}
Chun-Hao Chang, Rich Caruana, and Anna Goldenberg.
\newblock {NODE-GAM}: Neural generalized additive model for interpretable deep
  learning.
\newblock \emph{arXiv preprint arXiv:2106.01613}, 2021.

\bibitem[Chang et~al.(2017)Chang, Lu, and Chen]{chang2017aesthetic}
Kuang-Yu Chang, Kung-Hung Lu, and Chu-Song Chen.
\newblock Aesthetic critiques generation for photos.
\newblock In \emph{2017 IEEE International Conference on Computer Vision
  (ICCV)}, pp.\  3534--3543. IEEE, 2017.

\bibitem[Chen et~al.(2024)Chen, Ma, Song, Li, Hao, and Qin]{chen2024saliency}
Chenglizhao Chen, Guangxiao Ma, Wenfeng Song, Shuai Li, Aimin Hao, and Hong
  Qin.
\newblock Saliency-free and aesthetic-aware panoramic video navigation.
\newblock \emph{IEEE Transactions on Pattern Analysis and Machine
  Intelligence}, 47\penalty0 (3):\penalty0 2037--2054, 2024.

\bibitem[Chen et~al.(2016)Chen, Bai, Liang, and Li]{chen2016automatic}
Jiansheng Chen, Gaocheng Bai, Shaoheng Liang, and Zhengqin Li.
\newblock Automatic image cropping: A computational complexity study.
\newblock In \emph{Proceedings of the IEEE conference on computer vision and
  pattern recognition}, pp.\  507--515, 2016.

\bibitem[Chen et~al.(2018)Chen, Zhang, and Zhou]{chen2018fastgreedy}
Laming Chen, Guoxin Zhang, and Hanning Zhou.
\newblock Fast greedy {MAP} inference for determinantal point process to
  improve recommendation diversity.
\newblock In \emph{Advances in Neural Information Processing Systems 31,
  {NeurIPS} 2018}, 2018.

\bibitem[Chen et~al.(2017{\natexlab{a}})Chen, Huang, Chang, Tsai, Chen, and
  Chen]{chen2017quantitative}
Yi{-}Ling Chen, Tzu{-}Wei Huang, Kai{-}Han Chang, Yu{-}Chen Tsai, Hwann{-}Tzong
  Chen, and Bing{-}Yu Chen.
\newblock Quantitative analysis of automatic image cropping algorithms: {A}
  dataset and comparative study.
\newblock In \emph{2017 {IEEE} Winter Conference on Applications of Computer
  Vision, {WACV} 2017, Santa Rosa, CA, USA, March 24-31, 2017}, pp.\  226--234.
  {IEEE} Computer Society, 2017{\natexlab{a}}.
\newblock \doi{10.1109/WACV.2017.32}.

\bibitem[Chen et~al.(2017{\natexlab{b}})Chen, Klopp, Sun, Chien, and
  Ma]{chen2017compose}
Yi{-}Ling Chen, Jan Klopp, Min Sun, Shao{-}Yi Chien, and Kwan{-}Liu Ma.
\newblock Learning to compose with professional photographs on the web.
\newblock In \emph{Proceedings of the 2017 {ACM} on Multimedia Conference, {MM}
  2017, Mountain View, CA, USA, October 23-27, 2017}, pp.\  37--45. {ACM},
  2017{\natexlab{b}}.
\newblock \doi{10.1145/3123266.3123274}.

\bibitem[Dettmers et~al.(2023)Dettmers, Pagnoni, Holtzman, and
  Zettlemoyer]{dettmers2023qlora}
Tim Dettmers, Artidoro Pagnoni, Ari Holtzman, and Luke Zettlemoyer.
\newblock Qlora: Efficient finetuning of quantized llms.
\newblock In \emph{Advances in Neural Information Processing Systems 36: Annual
  Conference on Neural Information Processing Systems 2023, NeurIPS 2023, New
  Orleans, LA, USA, December 10 - 16, 2023}, 2023.
\newblock URL
  \url{http://papers.nips.cc/paper\_files/paper/2023/hash/1feb87871436031bdc0f2beaa62a049b-Abstract-Conference.html}.

\bibitem[Dong et~al.(2026)Dong, Li, Yu, and Chen]{dong2026crop}
Zhitong Dong, Chao Li, Jie Yu, and Hao Chen.
\newblock Crop: Expert-aligned image cropping via compositional reasoning and
  optimizing preference.
\newblock \emph{arXiv preprint arXiv:2605.12545}, 2026.

\bibitem[Du et~al.(2026)Du, He, and Peng]{du2026venus}
Tianxiang Du, Hulingxiao He, and Yuxin Peng.
\newblock Venus: Benchmarking and empowering multimodal large language models
  for aesthetic guidance and cropping.
\newblock \emph{CoRR}, abs/2602.23980, 2026.
\newblock \doi{10.48550/ARXIV.2602.23980}.

\bibitem[Fang et~al.(2014)Fang, Lin, M\v{e}ch, and Shen]{fang2014automatic}
Chen Fang, Zhe Lin, Radom{\'\i}r M\v{e}ch, and Xiaohui Shen.
\newblock Automatic image cropping using visual composition, boundary
  simplicity and content preservation models.
\newblock In \emph{Proceedings of the 22nd ACM International Conference on
  Multimedia}, MM '14, pp.\  1105--1108, New York, NY, USA, 2014. ACM.
\newblock \doi{10.1145/2647868.2654979}.

\bibitem[Fang \& Zhang(2017)Fang and Zhang]{fang2017creatism}
Hui Fang and Meng Zhang.
\newblock Creatism: {A} deep-learning photographer capable of creating
  professional work.
\newblock \emph{CoRR}, abs/1707.03491, 2017.
\newblock URL \url{http://arxiv.org/abs/1707.03491}.

\bibitem[Fang \& Chen(2025)Fang and Chen]{fang2025camerabench}
I-Sheng Fang and Jun-Cheng Chen.
\newblock {CameraBench}: Benchmarking visual reasoning in {MLLMs} via
  photography.
\newblock \emph{arXiv preprint arXiv:2504.10090}, 2025.

\bibitem[Feijoo et~al.(2025)Feijoo, Benito, Garcia, and
  Conde]{feijoo2025darkir}
Daniel Feijoo, Juan~C. Benito, Alvaro Garcia, and Marcos~V. Conde.
\newblock {DarkIR}: Robust low-light image restoration.
\newblock In \emph{2025 IEEE/CVF Conference on Computer Vision and Pattern
  Recognition (CVPR)}, pp.\  10879--10889. IEEE, 2025.

\bibitem[Gorishniy et~al.(2024)Gorishniy, Rubachev, Kartashev, Shlenskii,
  Kotelnikov, and Babenko]{gorishniy2024tabr}
Yury Gorishniy, Ivan Rubachev, Nikolay Kartashev, Daniil Shlenskii, Akim
  Kotelnikov, and Artem Babenko.
\newblock {TabR}: Tabular deep learning meets nearest neighbors.
\newblock In \emph{The Twelfth International Conference on Learning
  Representations, {ICLR} 2024}, 2024.

\bibitem[Grinsztajn et~al.(2022)Grinsztajn, Oyallon, and
  Varoquaux]{grinsztajn2022trees}
L{\'e}o Grinsztajn, Edouard Oyallon, and Ga{\"e}l Varoquaux.
\newblock Why do tree-based models still outperform deep learning on typical
  tabular data?
\newblock In \emph{Advances in Neural Information Processing Systems 35,
  Datasets and Benchmarks Track, {NeurIPS} 2022}, 2022.

\bibitem[Gu et~al.(2025)Gu, Pan, Hong, Liu, and Cao]{gu2025videoreframe}
Sitian Gu, Zhiyu Pan, Chaoyi Hong, Chengxin Liu, and Zhiguo Cao.
\newblock Dynamic beauty is easy to find: {A} large-scale composition-aware
  dataset and an end-to-end framework for video reframing.
\newblock In \emph{Proceedings of the 33rd {ACM} International Conference on
  Multimedia, {MM} 2025, Dublin, Ireland, October 27-31, 2025}, pp.\
  7776--7784. {ACM}, 2025.
\newblock \doi{10.1145/3746027.3755103}.

\bibitem[Guan et~al.(2024)Guan, Liu, Wu, Xian, Li, Liu, Wang, Chen, Huang,
  Yacoob, Manocha, and Zhou]{guan2024hallusionbench}
Tianrui Guan, Fuxiao Liu, Xiyang Wu, Ruiqi Xian, Zongxia Li, Xiaoyu Liu, Xijun
  Wang, Lichang Chen, Furong Huang, Yaser Yacoob, Dinesh Manocha, and Tianyi
  Zhou.
\newblock Hallusionbench: an advanced diagnostic suite for entangled language
  hallucination and visual illusion in large vision-language models.
\newblock In \emph{2024 IEEE/CVF Conference on Computer Vision and Pattern
  Recognition (CVPR)}, pp.\  14375--14385. IEEE, 2024.

\bibitem[Guo et~al.(2018)Guo, Wang, Shen, Yan, and Liao]{guo2018automatic}
Guanjun Guo, Hanzi Wang, Chunhua Shen, Yan Yan, and Hong-Yuan~Mark Liao.
\newblock Automatic image cropping for visual aesthetic enhancement using deep
  neural networks and cascaded regression.
\newblock \emph{IEEE Transactions on Multimedia}, 20\penalty0 (8):\penalty0
  2073--2085, 2018.

\bibitem[Hamara \& Freeman(2025)Hamara and Freeman]{hamara2025efficient}
Andrew Hamara and Andrew~C. Freeman.
\newblock Efficient multi-crop saliency partitioning for automatic image
  cropping.
\newblock In \emph{2025 International Conference on Modern Computing,
  Networking and Applications (MCNA)}, pp.\  105--108. IEEE, 2025.

\bibitem[He et~al.(2025)He, Tsai, Peng, Chen, Lin, and Lin]{he2026blurdm}
Jin-Ting He, Fu-Jen Tsai, Yan-Tsung Peng, Min-Hung Chen, Chia-Wen Lin, and
  Yen-Yu Lin.
\newblock {BlurDM}: A blur diffusion model for image deblurring.
\newblock In \emph{Advances in Neural Information Processing Systems},
  volume~38, pp.\  52196--52227, 2025.
\newblock \doi{10.52202/085713-1740}.

\bibitem[He et~al.(2022)He, Zhang, Xie, Jiang, and Ming]{he2022rethinking}
Shuai He, Yongchang Zhang, Rui Xie, Dongxiang Jiang, and Anlong Ming.
\newblock Rethinking image aesthetics assessment: Models, datasets and
  benchmarks.
\newblock In \emph{Proceedings of the Thirty-First International Joint
  Conference on Artificial Intelligence}, pp.\  942--948, 2022.

\bibitem[Holzm{\"u}ller et~al.(2024)Holzm{\"u}ller, Grinsztajn, and
  Steinwart]{holzmuller2024betterbydefault}
David Holzm{\"u}ller, L{\'e}o Grinsztajn, and Ingo Steinwart.
\newblock Better by default: Strong pre-tuned {MLP}s and boosted trees on
  tabular data.
\newblock In \emph{Advances in Neural Information Processing Systems 37,
  {NeurIPS} 2024}, 2024.

\bibitem[Hong et~al.(2021)Hong, Du, Xian, Lu, Cao, and Zhong]{hong2021cacnet}
Chaoyi Hong, Shuaiyuan Du, Ke~Xian, Hao Lu, Zhiguo Cao, and Weicai Zhong.
\newblock Composing photos like a photographer.
\newblock In \emph{{IEEE} Conference on Computer Vision and Pattern
  Recognition, {CVPR} 2021, virtual, June 19-25, 2021}, pp.\  7057--7066.
  Computer Vision Foundation / {IEEE}, 2021.
\newblock \doi{10.1109/CVPR46437.2021.00698}.

\bibitem[Hong et~al.(2024)Hong, Yuan, Gharbi, Fisher, and
  Fatahalian]{hong2024gencrop}
James Hong, Lu~Yuan, Micha{\"{e}}l Gharbi, Matthew Fisher, and Kayvon
  Fatahalian.
\newblock Learning subject-aware cropping by outpainting professional photos.
\newblock In \emph{Thirty-Eighth {AAAI} Conference on Artificial Intelligence,
  {AAAI} 2024, February 20-27, 2024, Vancouver, Canada}, pp.\  2175--2183.
  {AAAI} Press, 2024.
\newblock \doi{10.1609/aaai.v38i3.27990}.

\bibitem[Hsieh et~al.(2023)Hsieh, Li, Yeh, Nakhost, Fujii, Ratner, Krishna,
  Lee, and Pfister]{hsieh2023distilling}
Cheng-Yu Hsieh, Chun-Liang Li, Chih-kuan Yeh, Hootan Nakhost, Yasuhisa Fujii,
  Alex Ratner, Ranjay Krishna, Chen-Yu Lee, and Tomas Pfister.
\newblock Distilling step-by-step! outperforming larger language models with
  less training data and smaller model sizes.
\newblock In \emph{Findings of the association for computational linguistics:
  ACL 2023}, pp.\  8003--8017, 2023.

\bibitem[Hu et~al.(2022)Hu, Shen, Wallis, Allen{-}Zhu, Li, Wang, Wang, and
  Chen]{hu2022lora}
Edward~J. Hu, Yelong Shen, Phillip Wallis, Zeyuan Allen{-}Zhu, Yuanzhi Li,
  Shean Wang, Lu~Wang, and Weizhu Chen.
\newblock Lora: Low-rank adaptation of large language models.
\newblock In \emph{The Tenth International Conference on Learning
  Representations, {ICLR} 2022, Virtual Event, April 25-29, 2022}.
  OpenReview.net, 2022.
\newblock URL \url{https://openreview.net/forum?id=nZeVKeeFYf9}.

\bibitem[Hu et~al.(2017)Hu, Lin, Liu, Cheng, Chang, and Sun]{hu2017deep}
Hou-Ning Hu, Yen-Chen Lin, Ming-Yu Liu, Hsien-Tzu Cheng, Yung-Ju Chang, and Min
  Sun.
\newblock Deep 360 pilot: Learning a deep agent for piloting through
  360{\textdegree} sports videos.
\newblock In \emph{2017 IEEE Conference on Computer Vision and Pattern
  Recognition (CVPR)}, pp.\  1396--1405. IEEE, 2017.

\bibitem[Huang et~al.(2024{\natexlab{a}})Huang, Li, Chen, Wu, Lin, and
  Shi]{huang2024multi}
Yipo Huang, Leida Li, Pengfei Chen, Haoning Wu, Weisi Lin, and Guangming Shi.
\newblock Multi-modality multi-attribute contrastive pre-training for image
  aesthetics computing.
\newblock \emph{IEEE Transactions on Pattern Analysis and Machine
  Intelligence}, 47\penalty0 (2):\penalty0 1205--1218, 2024{\natexlab{a}}.

\bibitem[Huang et~al.(2024{\natexlab{b}})Huang, Sheng, Yang, Yuan, Duan, Chen,
  Li, Lin, and Shi]{huang2024aesexpert}
Yipo Huang, Xiangfei Sheng, Zhichao Yang, Quan Yuan, Zhichao Duan, Pengfei
  Chen, Leida Li, Weisi Lin, and Guangming Shi.
\newblock Aesexpert: Towards multi-modality foundation model for image
  aesthetics perception.
\newblock In \emph{Proceedings of the 32nd {ACM} International Conference on
  Multimedia, {MM} 2024, Melbourne, VIC, Australia, 28 October 2024 - 1
  November 2024}, pp.\  5911--5920. {ACM}, 2024{\natexlab{b}}.
\newblock \doi{10.1145/3664647.3680649}.

\bibitem[Huang et~al.(2024{\natexlab{c}})Huang, Yuan, Sheng, Yang, Wu, Chen,
  Yang, Li, and Lin]{huang2024aesbench}
Yipo Huang, Quan Yuan, Xiangfei Sheng, Zhichao Yang, Haoning Wu, Pengfei Chen,
  Yuzhe Yang, Leida Li, and Weisi Lin.
\newblock Aesbench: An expert benchmark for multimodal large language models on
  image aesthetics perception.
\newblock \emph{CoRR}, abs/2401.08276, 2024{\natexlab{c}}.
\newblock \doi{10.48550/ARXIV.2401.08276}.

\bibitem[Jacovi \& Goldberg(2020)Jacovi and Goldberg]{jacovi2020towards}
Alon Jacovi and Yoav Goldberg.
\newblock Towards faithfully interpretable nlp systems: How should we define
  and evaluate faithfulness?
\newblock In \emph{Proceedings of the 58th annual meeting of the association
  for computational linguistics}, pp.\  4198--4205, 2020.

\bibitem[Jia et~al.(2022)Jia, Huang, Fu, and He]{jia2022rethinking}
Gengyun Jia, Huaibo Huang, Chaoyou Fu, and Ran He.
\newblock Rethinking image cropping: Exploring diverse compositions from global
  views.
\newblock In \emph{{IEEE/CVF} Conference on Computer Vision and Pattern
  Recognition, {CVPR} 2022, New Orleans, LA, USA, June 18-24, 2022}, pp.\
  2436--2445. {IEEE}, 2022.
\newblock \doi{10.1109/CVPR52688.2022.00248}.

\bibitem[Jiang \& Chen(2025)Jiang and Chen]{jiang2025multimodal}
Ruixiang Jiang and Chang~Wen Chen.
\newblock Multimodal {LLMs} can reason about aesthetics in zero-shot.
\newblock In \emph{Proceedings of the 33rd ACM International Conference on
  Multimedia}, pp.\  6634--6643, 2025.

\bibitem[Jiang \& Chen(2026)Jiang and Chen]{jiang2026shutter}
Ruixiang Jiang and Chang~Wen Chen.
\newblock Before the shutter: Aesthetic and actionable portrait photography
  planning in 3d scenes.
\newblock \emph{CoRR}, abs/2605.30318, 2026.
\newblock \doi{10.48550/ARXIV.2605.30318}.

\bibitem[Kang \& Cho(2019)Kang and Cho]{kang2019interactive}
Kyoungkook Kang and Sunghyun Cho.
\newblock Interactive and automatic navigation for 360{\textdegree} video
  playback.
\newblock \emph{ACM Transactions on Graphics (TOG)}, 38\penalty0 (4):\penalty0
  1--11, 2019.

\bibitem[Ke et~al.(2017)Ke, Meng, Finley, Wang, Chen, Ma, Ye, and
  Liu]{ke2017lightgbm}
Guolin Ke, Qi~Meng, Thomas Finley, Taifeng Wang, Wei Chen, Weidong Ma, Qiwei
  Ye, and Tie{-}Yan Liu.
\newblock Lightgbm: {A} highly efficient gradient boosting decision tree.
\newblock In \emph{Advances in Neural Information Processing Systems 30: Annual
  Conference on Neural Information Processing Systems 2017, December 4-9, 2017,
  Long Beach, CA, {USA}}, pp.\  3146--3154, 2017.
\newblock URL
  \url{https://proceedings.neurips.cc/paper/2017/hash/6449f44a102fde848669bdd9eb6b76fa-Abstract.html}.

\bibitem[Ke et~al.(2023)Ke, Ye, Yu, Wu, Milanfar, and Yang]{ke2023vila}
Junjie Ke, Keren Ye, Jiahui Yu, Yonghui Wu, Peyman Milanfar, and Feng Yang.
\newblock {VILA}: Learning image aesthetics from user comments with
  vision-language pretraining.
\newblock In \emph{2023 IEEE/CVF Conference on Computer Vision and Pattern
  Recognition (CVPR)}, pp.\  10041--10051. IEEE, 2023.

\bibitem[Kim et~al.(2025)Kim, Afifi, Kim, Brown, and Kim]{kim2025ccmnet}
Dongyoung Kim, Mahmoud Afifi, Dongyun Kim, Michael~S. Brown, and Seon~Joo Kim.
\newblock {CCMNet}: Leveraging calibrated color correction matrices for
  cross-camera color constancy.
\newblock In \emph{2025 IEEE/CVF International Conference on Computer Vision
  (ICCV)}, pp.\  6198--6208. IEEE, 2025.

\bibitem[Kim et~al.(2022)Kim, Kim, Lee, Cha, Lee, and Kim]{kim2022inspyrenet}
Taehun Kim, Kunhee Kim, Joonyeong Lee, Dongmin Cha, Jiho Lee, and Daijin Kim.
\newblock Revisiting image pyramid structure for high resolution salient object
  detection.
\newblock In \emph{Computer Vision - {ACCV} 2022 - 16th Asian Conference on
  Computer Vision, Macao, China, December 4-8, 2022}, Lecture Notes in Computer
  Science. Springer, 2022.
\newblock \doi{10.1007/978-3-031-26293-7_16}.

\bibitem[Kim et~al.(2017)Kim, Lee, Cho, Kwon, Choi, and Yoon]{kim2017automatic}
Yeong~Won Kim, Chang-Ryeol Lee, Dae-Yong Cho, Yong~Hoon Kwon, Hyeok-Jae Choi,
  and Kuk-Jin Yoon.
\newblock Automatic content-aware projection for 360{\textdegree} videos.
\newblock In \emph{2017 IEEE International Conference on Computer Vision
  (ICCV)}, pp.\  4753--4761. IEEE, 2017.

\bibitem[Kirillov et~al.(2023)Kirillov, Mintun, Ravi, Mao, Rolland, Gustafson,
  Xiao, Whitehead, Berg, Lo, Doll{\'{a}}r, and Girshick]{kirillov2023sam}
Alexander Kirillov, Eric Mintun, Nikhila Ravi, Hanzi Mao, Chloe Rolland, Laura
  Gustafson, Tete Xiao, Spencer Whitehead, Alexander~C. Berg, Wan{-}Yen Lo,
  Piotr Doll{\'{a}}r, and Ross Girshick.
\newblock Segment anything.
\newblock In \emph{{IEEE/CVF} International Conference on Computer Vision,
  {ICCV} 2023, Paris, France, October 1-6, 2023}, pp.\  3992--4003. {IEEE},
  2023.
\newblock \doi{10.1109/ICCV51070.2023.00371}.

\bibitem[Kong et~al.(2025)Kong, Zhang, Zou, Wang, Ren, Wu, Dong, and
  Pan]{kong2026deblurdiff}
Lingshun Kong, Jiawei Zhang, Dongqing Zou, Fu~Lee Wang, Jimmy~S. Ren, Xiaohe
  Wu, Jiangxin Dong, and Jinshan Pan.
\newblock {DeblurDiff}: Real-world image deblurring with generative diffusion
  models.
\newblock In \emph{Advances in Neural Information Processing Systems},
  volume~38, pp.\  154855--154873, 2025.
\newblock \doi{10.52202/085713-5179}.

\bibitem[Kulesza \& Taskar(2011)Kulesza and Taskar]{kulesza2011kdpp}
Alex Kulesza and Ben Taskar.
\newblock k-{DPP}s: Fixed-size determinantal point processes.
\newblock In \emph{Proceedings of the 28th International Conference on Machine
  Learning, {ICML} 2011}, 2011.

\bibitem[Kulesza \& Taskar(2012)Kulesza and Taskar]{kulesza2012dpp}
Alex Kulesza and Ben Taskar.
\newblock Determinantal point processes for machine learning.
\newblock \emph{Foundations and Trends in Machine Learning}, 5\penalty0
  (2-3):\penalty0 123--286, 2012.
\newblock \doi{10.1561/2200000044}.

\bibitem[Lai et~al.(2018)Lai, Huang, Joshi, Buehler, Yang, and
  Kang]{lai2017semantic}
Wei-Sheng Lai, Yujia Huang, Neel Joshi, Christopher Buehler, Ming-Hsuan Yang,
  and Sing~Bing Kang.
\newblock Semantic-driven generation of hyperlapse from 360 degree video.
\newblock \emph{IEEE transactions on visualization and computer graphics},
  24\penalty0 (9):\penalty0 2610--2621, 2018.

\bibitem[Lee et~al.(2018{\natexlab{a}})Lee, Kim, Lee, and
  Kim]{lee2018photographic}
Jun-Tae Lee, Han-Ul Kim, Chul Lee, and Chang-Su Kim.
\newblock Photographic composition classification and dominant geometric
  element detection for outdoor scenes.
\newblock \emph{Journal of Visual Communication and Image Representation},
  55:\penalty0 91--105, 2018{\natexlab{a}}.

\bibitem[Lee et~al.(2024)Lee, Shin, and Lee]{lee2024learning}
Kyunghyun Lee, Ukcheol Shin, and Byeong-Uk Lee.
\newblock Learning to control camera exposure via reinforcement learning.
\newblock In \emph{2024 IEEE/CVF Conference on Computer Vision and Pattern
  Recognition (CVPR)}, pp.\  2975--2983. IEEE, 2024.

\bibitem[Lee et~al.(2018{\natexlab{b}})Lee, Sung, Yu, and Kim]{lee2018memory}
Sangho Lee, Jinyoung Sung, Youngjae Yu, and Gunhee Kim.
\newblock A memory network approach for story-based temporal summarization of
  360{\textdegree} videos.
\newblock In \emph{2018 IEEE/CVF Conference on Computer Vision and Pattern
  Recognition}, pp.\  1410--1419. IEEE, 2018{\natexlab{b}}.

\bibitem[Lee et~al.(2025)Lee, Jiang, Xu, Li, Ke, Li, He, Hickson, Datsenko,
  Kim, Yang, Essa, and Yang]{lee2025cropper}
Seung~Hyun Lee, Jijun Jiang, Yiran Xu, Zhuofang Li, Junjie Ke, Yinxiao Li,
  Junfeng He, Steven Hickson, Katie Datsenko, Sangpil Kim, Ming{-}Hsuan Yang,
  Irfan Essa, and Feng Yang.
\newblock Cropper: Vision-language model for image cropping through in-context
  learning.
\newblock In \emph{{IEEE/CVF} Conference on Computer Vision and Pattern
  Recognition, {CVPR} 2025, Nashville, TN, USA, June 11-15, 2025}, pp.\
  30010--30019. Computer Vision Foundation / {IEEE}, 2025.
\newblock \doi{10.1109/CVPR52734.2025.02793}.

\bibitem[Lee \& Ding(2024)Lee and Ding]{lee2024efficient}
Yuan-Kang Lee and Jian-Jiun Ding.
\newblock Efficient color image denoising using dwt-based noise estimation and
  adaptive wiener filter.
\newblock In \emph{2024 8th International Conference on Imaging, Signal
  Processing and Communications (ICISPC)}, pp.\  47--51. IEEE, 2024.

\bibitem[Lee et~al.(2026{\natexlab{a}})Lee, Chen, Chang, and Liu]{lee2026rl}
Yuan-Kang Lee, Kuan-Lin Chen, Chia-Che Chang, and Yu-Lun Liu.
\newblock Rl-awb: Deep reinforcement learning for auto white balance correction
  in low-light nighttime scenes.
\newblock In \emph{European Conference on Computer Vision}, pp.\  77--99.
  Springer, 2026{\natexlab{a}}.

\bibitem[Lee et~al.(2026{\natexlab{b}})Lee, Chen, Chang, and Ding]{lee2026safe}
Yuan-Kang Lee, Kuan-Lin Chen, Chih-Heng Chang, and Jian-Jiun Ding.
\newblock Safe: Scene-aware feature modulation for color constancy with learned
  color space in pure-color scenes.
\newblock \emph{arXiv preprint arXiv:2608.13967}, 2026{\natexlab{b}}.

\bibitem[Li et~al.(2018)Li, Wu, Zhang, and Huang]{li2018a2rl}
Debang Li, Huikai Wu, Junge Zhang, and Kaiqi Huang.
\newblock {A2-RL:} aesthetics aware reinforcement learning for image cropping.
\newblock In \emph{2018 {IEEE} Conference on Computer Vision and Pattern
  Recognition, {CVPR} 2018, Salt Lake City, UT, USA, June 18-22, 2018}, pp.\
  8193--8201. Computer Vision Foundation / {IEEE} Computer Society, 2018.
\newblock \doi{10.1109/CVPR.2018.00855}.

\bibitem[Li et~al.(2019)Li, Wu, Zhang, and Huang]{li2019a3rl}
Debang Li, Huikai Wu, Junge Zhang, and Kaiqi Huang.
\newblock Fast {A3RL:} aesthetics-aware adversarial reinforcement learning for
  image cropping.
\newblock \emph{{IEEE} Trans. Image Process.}, 28\penalty0 (10):\penalty0
  5105--5120, 2019.
\newblock \doi{10.1109/TIP.2019.2914360}.

\bibitem[Li et~al.(2020{\natexlab{a}})Li, Zhang, and Huang]{li2020mars}
Debang Li, Junge Zhang, and Kaiqi Huang.
\newblock Learning to learn cropping models for different aspect ratio
  requirements.
\newblock In \emph{2020 {IEEE/CVF} Conference on Computer Vision and Pattern
  Recognition, {CVPR} 2020, Seattle, WA, USA, June 13-19, 2020}, pp.\
  12682--12691. Computer Vision Foundation / {IEEE}, 2020{\natexlab{a}}.
\newblock \doi{10.1109/CVPR42600.2020.01270}.

\bibitem[Li et~al.(2020{\natexlab{b}})Li, Zhang, Huang, and Yang]{li2020cgs}
Debang Li, Junge Zhang, Kaiqi Huang, and Ming{-}Hsuan Yang.
\newblock Composing good shots by exploiting mutual relations.
\newblock In \emph{2020 {IEEE/CVF} Conference on Computer Vision and Pattern
  Recognition, {CVPR} 2020, Seattle, WA, USA, June 13-19, 2020}, pp.\
  4212--4221. Computer Vision Foundation / {IEEE}, 2020{\natexlab{b}}.
\newblock \doi{10.1109/CVPR42600.2020.00427}.

\bibitem[Li et~al.(2025{\natexlab{a}})Li, Zhou, Zhong, Lin, and
  Qiu]{li2025spas}
Jiawan Li, Fei Zhou, Zhipeng Zhong, Jiongzhi Lin, and Guoping Qiu.
\newblock Towards smart point-and-shoot photography.
\newblock In \emph{{IEEE/CVF} Conference on Computer Vision and Pattern
  Recognition, {CVPR} 2025, Nashville, TN, USA, June 11-15, 2025}, pp.\
  28242--28251. Computer Vision Foundation / {IEEE}, 2025{\natexlab{a}}.
\newblock \doi{10.1109/CVPR52734.2025.02630}.

\bibitem[Li et~al.(2026)Li, Fang, Hu, Cheng, Huang, Fan, Yu, and
  Ma]{li2026shuttermuse}
Jiayu Li, Yixiao Fang, Tianyu Hu, Wei Cheng, Ping Huang, Zheheng Fan, Gang Yu,
  and Xingjun Ma.
\newblock Shuttermuse: Capture-time photography guidance with mllms.
\newblock \emph{arXiv preprint arXiv:2606.25763}, 2026.

\bibitem[Li et~al.(2025{\natexlab{b}})Li, Zhang, Zhao, Zhang, Li, Zhang, and
  Zhang]{li2026q}
Weiqi Li, Xuanyu Zhang, Shijie Zhao, Yabin Zhang, Junlin Li, Li~Zhang, and Jian
  Zhang.
\newblock {Q-Insight}: Understanding image quality via visual reinforcement
  learning.
\newblock In \emph{Advances in Neural Information Processing Systems},
  volume~38, pp.\  36802--36827, 2025{\natexlab{b}}.

\bibitem[Li et~al.(2023)Li, Du, Zhou, Wang, Zhao, and Wen]{li2023evaluating}
Yifan Li, Yifan Du, Kun Zhou, Jinpeng Wang, Xin Zhao, and Ji-Rong Wen.
\newblock Evaluating object hallucination in large vision-language models.
\newblock In \emph{Proceedings of the 2023 conference on empirical methods in
  natural language processing}, pp.\  292--305, 2023.

\bibitem[Limoyo et~al.(2024)Limoyo, Li, Rivkin, Kelly, and
  Dudek]{limoyo2024photobot}
Oliver Limoyo, Jimmy Li, Dmitriy Rivkin, Jonathan Kelly, and Gregory Dudek.
\newblock Photobot: Reference-guided interactive photography via natural
  language.
\newblock In \emph{2024 IEEE/RSJ International Conference on Intelligent Robots
  and Systems (IROS)}, pp.\  2479--2486. IEEE, 2024.

\bibitem[Liu et~al.(2010)Liu, Chen, Wolf, and Cohen-Or]{liu2010optimizing}
Ligang Liu, Renjie Chen, Lior Wolf, and Daniel Cohen-Or.
\newblock Optimizing photo composition.
\newblock In \emph{Computer graphics forum}, volume~29, pp.\  469--478. Wiley
  Online Library, 2010.

\bibitem[Liu \& Wagemans(2026)Liu and Wagemans]{liu2026concepts}
Xiao{-}Chang Liu and Johan Wagemans.
\newblock From concepts to judgments: Interpretable image aesthetic assessment.
\newblock \emph{CoRR}, abs/2603.18108, 2026.
\newblock \doi{10.48550/ARXIV.2603.18108}.

\bibitem[Liu et~al.(2023)Liu, Liu, Li, Liu, Wang, Lei, and Zuo]{liu2023unic}
Xiaoyu Liu, Ming Liu, Junyi Li, Shuai Liu, Xiaotao Wang, Lei Lei, and Wangmeng
  Zuo.
\newblock Beyond image borders: Learning feature extrapolation for unbounded
  image composition.
\newblock In \emph{{IEEE/CVF} International Conference on Computer Vision,
  {ICCV} 2023, Paris, France, October 1-6, 2023}, pp.\  12977--12986. {IEEE},
  2023.
\newblock \doi{10.1109/ICCV51070.2023.01197}.

\bibitem[Lou et~al.(2013)Lou, Caruana, Gehrke, and Hooker]{lou2013accurate}
Yin Lou, Rich Caruana, Johannes Gehrke, and Giles Hooker.
\newblock Accurate intelligible models with pairwise interactions.
\newblock In \emph{Proceedings of the 19th ACM SIGKDD international conference
  on Knowledge discovery and data mining}, pp.\  623--631, 2013.

\bibitem[Lu et~al.(2019)Lu, Xing, Cai, and Xu]{lu2019listwise}
Weirui Lu, Xiaofen Xing, Bolun Cai, and Xiangmin Xu.
\newblock Listwise view ranking for image cropping.
\newblock \emph{{IEEE} Access}, 7:\penalty0 91904--91911, 2019.
\newblock \doi{10.1109/ACCESS.2019.2925430}.

\bibitem[Lundberg \& Lee(2017)Lundberg and Lee]{lundberg2017shap}
Scott~M. Lundberg and Su{-}In Lee.
\newblock A unified approach to interpreting model predictions.
\newblock In \emph{Advances in Neural Information Processing Systems 30: Annual
  Conference on Neural Information Processing Systems 2017, December 4-9, 2017,
  Long Beach, CA, {USA}}, pp.\  4765--4774, 2017.
\newblock URL
  \url{https://proceedings.neurips.cc/paper/2017/hash/8a20a8621978632d76c43dfd28b67767-Abstract.html}.

\bibitem[Lundberg et~al.(2020)Lundberg, Erion, Chen, DeGrave, Prutkin, Nair,
  Katz, Himmelfarb, Bansal, and Lee]{lundberg2020treeshap}
Scott~M. Lundberg, Gabriel Erion, Hugh Chen, Alex DeGrave, Jordan~M. Prutkin,
  Bala Nair, Ronit Katz, Jonathan Himmelfarb, Nisha Bansal, and Su{-}In Lee.
\newblock From local explanations to global understanding with explainable {AI}
  for trees.
\newblock \emph{Nature Machine Intelligence}, 2\penalty0 (1):\penalty0 56--67,
  2020.
\newblock \doi{10.1038/s42256-019-0138-9}.

\bibitem[McElfresh et~al.(2023)McElfresh, Khandagale, Valverde, C,
  Ramakrishnan, Goldblum, and White]{mcelfresh2023neuralvstrees}
Duncan McElfresh, Sujay Khandagale, Jonathan Valverde, Vishak~Prasad C, Ganesh
  Ramakrishnan, Micah Goldblum, and Colin White.
\newblock When do neural nets outperform boosted trees on tabular data?
\newblock In \emph{Advances in Neural Information Processing Systems 36,
  Datasets and Benchmarks Track, {NeurIPS} 2023}, 2023.

\bibitem[Mohammadi et~al.(2026)Mohammadi, Honari, Tsogkas, Aumentado-Armstrong,
  Brown, Mohomed, Derpanis, Levinshtein, and Gilitschenski]{mohammadi2026low}
Mohammad Mohammadi, Sina Honari, Stavros Tsogkas, Tristan Aumentado-Armstrong,
  Michael~S. Brown, Iqbal Mohomed, Konstantinos~G. Derpanis, Alex Levinshtein,
  and Igor Gilitschenski.
\newblock Why low-light cameras go color blind: Removing color bias in raw
  denoising.
\newblock \emph{arXiv preprint arXiv:2607.11090}, 2026.

\bibitem[Murray et~al.(2012)Murray, Marchesotti, and Perronnin]{murray2012ava}
Naila Murray, Luca Marchesotti, and Florent Perronnin.
\newblock {AVA:} {A} large-scale database for aesthetic visual analysis.
\newblock In \emph{2012 {IEEE} Conference on Computer Vision and Pattern
  Recognition, Providence, RI, USA, June 16-21, 2012}, pp.\  2408--2415. {IEEE}
  Computer Society, 2012.
\newblock \doi{10.1109/CVPR.2012.6247954}.

\bibitem[Niculescu-Mizil \& Caruana(2005)Niculescu-Mizil and
  Caruana]{niculescu2005predicting}
Alexandru Niculescu-Mizil and Rich Caruana.
\newblock Predicting good probabilities with supervised learning.
\newblock In \emph{Proceedings of the 22nd international conference on Machine
  learning}, pp.\  625--632, 2005.

\bibitem[Nishiyama et~al.(2009)Nishiyama, Okabe, Sato, and
  Sato]{nishiyama2009sensation}
Masashi Nishiyama, Takahiro Okabe, Yoichi Sato, and Imari Sato.
\newblock Sensation-based photo cropping.
\newblock In \emph{Proceedings of the 17th ACM international conference on
  Multimedia}, pp.\  669--672, 2009.

\bibitem[Nori et~al.(2019)Nori, Jenkins, Koch, and
  Caruana]{nori2019interpretml}
Harsha Nori, Samuel Jenkins, Paul Koch, and Rich Caruana.
\newblock {InterpretML}: A unified framework for machine learning
  interpretability.
\newblock \emph{arXiv preprint arXiv:1909.09223}, 2019.

\bibitem[Pan et~al.(2021)Pan, Cao, Wang, Lu, and Zhong]{pan2021transview}
Zhiyu Pan, Zhiguo Cao, Kewei Wang, Hao Lu, and Weicai Zhong.
\newblock Transview: Inside, outside, and across the cropping view boundaries.
\newblock In \emph{2021 {IEEE/CVF} International Conference on Computer Vision,
  {ICCV} 2021, Montreal, QC, Canada, October 10-17, 2021}, pp.\  4198--4207.
  {IEEE}, 2021.
\newblock \doi{10.1109/ICCV48922.2021.00418}.

\bibitem[Pan et~al.(2023)Pan, Chen, Zhang, Lu, Cao, and Zhong]{pan2023c2c}
Zhiyu Pan, Yinpeng Chen, Jiale Zhang, Hao Lu, Zhiguo Cao, and Weicai Zhong.
\newblock Find beauty in the rare: Contrastive composition feature clustering
  for nontrivial cropping box regression.
\newblock In \emph{Thirty-Seventh {AAAI} Conference on Artificial Intelligence,
  {AAAI} 2023, Washington, DC, USA, February 7-14, 2023}, pp.\  2011--2019.
  {AAAI} Press, 2023.
\newblock \doi{10.1609/aaai.v37i2.25293}.

\bibitem[Pan et~al.(2024)Pan, Cui, Wang, Wu, and Cao]{pan2024pseudo}
Zhiyu Pan, Jiahao Cui, Kewei Wang, Yizheng Wu, and Zhiguo Cao.
\newblock Pseudo label fusion with uncertainty estimation for semi-supervised
  cropping box regression.
\newblock \emph{{IEEE} Trans. Multim.}, 26:\penalty0 8157--8171, 2024.
\newblock \doi{10.1109/TMM.2024.3377125}.

\bibitem[Qi et~al.(2025)Qi, Zhao, Shi, Jenni, Fan, Dernoncourt, Cohen, and
  Li]{qi2025photographerseye}
Daiqing Qi, Handong Zhao, Jing Shi, Simon Jenni, Yifei Fan, Franck Dernoncourt,
  Scott Cohen, and Sheng Li.
\newblock The photographer's eye: Teaching multimodal large language models to
  see, and critique like photographers.
\newblock In \emph{{IEEE/CVF} Conference on Computer Vision and Pattern
  Recognition, {CVPR} 2025, Nashville, TN, USA, June 11-15, 2025}, pp.\
  24807--24816. Computer Vision Foundation / {IEEE}, 2025.
\newblock \doi{10.1109/CVPR52734.2025.02310}.

\bibitem[Radenovic et~al.(2022)Radenovic, Dubey, and
  Mahajan]{radenovic2022neural}
Filip Radenovic, Abhimanyu Dubey, and Dhruv Mahajan.
\newblock Neural basis models for interpretability.
\newblock \emph{Advances in Neural Information Processing Systems},
  35:\penalty0 8414--8426, 2022.

\bibitem[Radford et~al.(2021)Radford, Kim, Hallacy, Ramesh, Goh, Agarwal,
  Sastry, Askell, Mishkin, Clark, Krueger, and Sutskever]{radford2021clip}
Alec Radford, Jong~Wook Kim, Chris Hallacy, Aditya Ramesh, Gabriel Goh,
  Sandhini Agarwal, Girish Sastry, Amanda Askell, Pamela Mishkin, Jack Clark,
  Gretchen Krueger, and Ilya Sutskever.
\newblock Learning transferable visual models from natural language
  supervision.
\newblock In \emph{Proceedings of the 38th International Conference on Machine
  Learning, {ICML} 2021, 18-24 July 2021, Virtual Event}, volume 139 of
  \emph{Proceedings of Machine Learning Research}, pp.\  8748--8763. {PMLR},
  2021.

\bibitem[Rai et~al.(2017)Rai, Guti{\'e}rrez, and Le~Callet]{rai2017dataset}
Yashas Rai, Jes{\'u}s Guti{\'e}rrez, and Patrick Le~Callet.
\newblock A dataset of head and eye movements for 360 degree images.
\newblock In \emph{Proceedings of the 8th ACM on Multimedia Systems
  Conference}, pp.\  205--210, 2017.

\bibitem[Rao et~al.(2022)Rao, Xu, and Lin]{rao2022shoot360}
Anyi Rao, Linning Xu, and Dahua Lin.
\newblock Shoot360: Normal view video creation from city panorama footage.
\newblock In \emph{{SIGGRAPH} '22: Special Interest Group on Computer Graphics
  and Interactive Techniques Conference, Vancouver, BC, Canada, August 7 - 11,
  2022}, pp.\  13:1--13:9. {ACM}, 2022.
\newblock \doi{10.1145/3528233.3530702}.

\bibitem[Schuhmann et~al.(2022)Schuhmann, Beaumont, Vencu, Gordon, Wightman,
  Cherti, Coombes, Katta, Mullis, Wortsman, Schramowski, Kundurthy, Crowson,
  Schmidt, Kaczmarczyk, and Jitsev]{schuhmann2022laion5b}
Christoph Schuhmann, Romain Beaumont, Richard Vencu, Cade Gordon, Ross
  Wightman, Mehdi Cherti, Theo Coombes, Aarush Katta, Clayton Mullis, Mitchell
  Wortsman, Patrick Schramowski, Srivatsa Kundurthy, Katherine Crowson, Ludwig
  Schmidt, Robert Kaczmarczyk, and Jenia Jitsev.
\newblock {LAION-5B}: An open large-scale dataset for training next generation
  image-text models.
\newblock In \emph{Advances in Neural Information Processing Systems 35:
  Datasets and Benchmarks Track}, 2022.
\newblock \doi{10.52202/068431-1833}.

\bibitem[Sheng et~al.(2025)Sheng, Xie, Zou, Chen, Zhu, and
  Li]{sheng2025instructcrop}
Xiangfei Sheng, Pangu Xie, Weidong Zou, Pengfei Chen, Tong Zhu, and Leida Li.
\newblock Instructcrop: Teaching multimodal large language models to crop
  aesthetic images.
\newblock In \emph{Proceedings of the 33rd ACM International Conference on
  Multimedia}, pp.\  6830--6839, 2025.
\newblock \doi{10.1145/3746027.3754931}.
\newblock URL \url{https://dl.acm.org/doi/10.1145/3746027.3754931}.

\bibitem[Shin et~al.(2024)Shin, Lee, Ko, and Kim]{shin2024order}
Nyeong{-}Ho Shin, Seon{-}Ho Lee, Jinwon Ko, and Chang{-}Su Kim.
\newblock Image cropping based on order learning.
\newblock \emph{J. Vis. Commun. Image Represent.}, 103:\penalty0 104253, 2024.
\newblock \doi{10.1016/J.JVCIR.2024.104253}.

\bibitem[Sim{\'{e}}oni et~al.(2026)Sim{\'{e}}oni, Vo, Seitzer, Baldassarre,
  Oquab, Jose, Khalidov, Szafraniec, Yi, Ramamonjisoa, Massa, Haziza,
  Wehrstedt, Wang, Darcet, Moutakanni, Sentana, Roberts, Vedaldi, Tolan,
  Brandt, Couprie, Mairal, J{\'{e}}gou, Labatut, and
  Bojanowski]{simeoni2026dinov3}
Oriane Sim{\'{e}}oni, Huy~V. Vo, Maximilian Seitzer, Federico Baldassarre,
  Maxime Oquab, Cijo Jose, Vasil Khalidov, Marc Szafraniec, Seung~Eun Yi,
  Micha{\"{e}}l Ramamonjisoa, Francisco Massa, Daniel Haziza, Luca Wehrstedt,
  Jianyuan Wang, Timoth{\'{e}}e Darcet, Th{\'{e}}o Moutakanni, Leonel Sentana,
  Claire Roberts, Andrea Vedaldi, Jamie Tolan, John Brandt, Camille Couprie,
  Julien Mairal, Herv{\'{e}} J{\'{e}}gou, Patrick Labatut, and Piotr
  Bojanowski.
\newblock Dinov3.
\newblock \emph{Trans. Mach. Learn. Res.}, 2026, 2026.
\newblock URL \url{https://openreview.net/forum?id=2NlGyqNjns}.

\bibitem[Sitzmann et~al.(2018)Sitzmann, Serrano, Pavel, Agrawala,
  Guti{\'e}rrez, Masi{\'a}, and Wetzstein]{sitzmann2018saliency}
Vincent Sitzmann, Ana Serrano, Amy Pavel, Maneesh Agrawala, Diego
  Guti{\'e}rrez, Bel{\'e}n Masi{\'a}, and Gordon Wetzstein.
\newblock Saliency in vr: How do people explore virtual environments?
\newblock \emph{IEEE transactions on visualization and computer graphics},
  24\penalty0 (4):\penalty0 1633--1642, 2018.

\bibitem[Su \& Grauman(2017)Su and Grauman]{su2017making}
Yu-Chuan Su and Kristen Grauman.
\newblock Making 360{\textdegree} video watchable in {2D}: Learning videography
  for click free viewing.
\newblock In \emph{2017 IEEE Conference on Computer Vision and Pattern
  Recognition (CVPR)}, pp.\  1368--1376. IEEE, 2017.

\bibitem[Su et~al.(2016)Su, Jayaraman, and Grauman]{su2016pano2vid}
Yu{-}Chuan Su, Dinesh Jayaraman, and Kristen Grauman.
\newblock Pano2vid: Automatic cinematography for watching 360{\textdegree}
  videos.
\newblock In \emph{Computer Vision - {ACCV} 2016 - 13th Asian Conference on
  Computer Vision, Taipei, Taiwan, November 20-24, 2016, Revised Selected
  Papers, Part {IV}}, volume 10114 of \emph{Lecture Notes in Computer Science},
  pp.\  154--171. Springer, 2016.
\newblock \doi{10.1007/978-3-319-54190-7\_10}.

\bibitem[Su et~al.(2021)Su, Vemulapalli, Weiss, Chu, Mansfield, Shapira, and
  Pitts]{su2021viewadjust}
Yu{-}Chuan Su, Raviteja Vemulapalli, Ben Weiss, Chun{-}Te Chu, Philip~Andrew
  Mansfield, Lior Shapira, and Colvin Pitts.
\newblock Camera view adjustment prediction for improving image composition.
\newblock \emph{CoRR}, abs/2104.07608, 2021.
\newblock URL \url{https://arxiv.org/abs/2104.07608}.

\bibitem[Su et~al.(2024)Su, Cao, Deng, Rao, and Wu]{su2024s2cnet}
Yukun Su, Yiwen Cao, Jingliang Deng, Fengyun Rao, and Qingyao Wu.
\newblock Spatial-semantic collaborative cropping for user generated content.
\newblock In \emph{Thirty-Eighth {AAAI} Conference on Artificial Intelligence,
  {AAAI} 2024, February 20-27, 2024, Vancouver, Canada}, pp.\  4988--4997.
  {AAAI} Press, 2024.
\newblock \doi{10.1609/aaai.v38i5.28303}.

\bibitem[Talebi \& Milanfar(2018)Talebi and Milanfar]{talebi2018nima}
Hossein Talebi and Peyman Milanfar.
\newblock {NIMA:} neural image assessment.
\newblock \emph{{IEEE} Trans. Image Process.}, 27\penalty0 (8):\penalty0
  3998--4011, 2018.
\newblock \doi{10.1109/TIP.2018.2831899}.

\bibitem[Tang et~al.(2026)Tang, Sarvestani, Xu, Xu, and
  Wang]{tang2026aestheticfield}
Sheyang Tang, Armin~Shafiee Sarvestani, Jialu Xu, Xiaoyu Xu, and Zhou Wang.
\newblock Aesthetic camera viewpoint suggestion with 3d aesthetic field.
\newblock \emph{CoRR}, abs/2602.20363, 2026.
\newblock \doi{10.48550/ARXIV.2602.20363}.

\bibitem[Tu et~al.(2020)Tu, Niu, Zhao, Cheng, and Zhang]{tu2020asm}
Yi~Tu, Li~Niu, Weijie Zhao, Dawei Cheng, and Liqing Zhang.
\newblock Image cropping with composition and saliency aware aesthetic score
  map.
\newblock In \emph{The Thirty-Fourth {AAAI} Conference on Artificial
  Intelligence, {AAAI} 2020, New York, NY, USA, February 7-12, 2020}, pp.\
  12104--12111. {AAAI} Press, 2020.
\newblock \doi{10.1609/aaai.v34i07.6889}.

\bibitem[Uchida et~al.(2025)Uchida, Kanamori, and Endo]{uchida20253dview}
Taichi Uchida, Yoshihiro Kanamori, and Yuki Endo.
\newblock 3d view optimization for improving image aesthetics.
\newblock In \emph{2025 {IEEE} International Conference on Acoustics, Speech
  and Signal Processing, {ICASSP} 2025, Hyderabad, India, April 6-11, 2025},
  pp.\  1--5. {IEEE}, 2025.
\newblock \doi{10.1109/ICASSP49660.2025.10888966}.

\bibitem[Viriyavisuthisakul et~al.(2025)Viriyavisuthisakul, Yoshida, Shiohara,
  Xiao, and Yamasaki]{viriyavisuthisakul2025xai}
Supatta Viriyavisuthisakul, Shun Yoshida, Kaede Shiohara, Ling Xiao, and
  Toshihiko Yamasaki.
\newblock Explainable {AI} for image aesthetic evaluation using vision-language
  models.
\newblock In \emph{2025 Conference on Artificial Intelligence x Multimedia
  {(AIxMM)}}, pp.\  62--65. {IEEE}, 2025.
\newblock \doi{10.1109/aixmm62960.2025.00017}.

\bibitem[Wang et~al.(2023{\natexlab{a}})Wang, Niu, Zhang, and
  Zhang]{wang2023spatial}
Chao Wang, Li~Niu, Bo~Zhang, and Liqing Zhang.
\newblock Image cropping with spatial-aware feature and rank consistency.
\newblock In \emph{{IEEE/CVF} Conference on Computer Vision and Pattern
  Recognition, {CVPR} 2023, Vancouver, BC, Canada, June 17-24, 2023}, pp.\
  10052--10061. {IEEE}, 2023{\natexlab{a}}.
\newblock \doi{10.1109/CVPR52729.2023.00969}.

\bibitem[Wang et~al.(2023{\natexlab{b}})Wang, Chan, and Loy]{wang2023exploring}
Jianyi Wang, Kelvin C.~K. Chan, and Chen~Change Loy.
\newblock Exploring {CLIP} for assessing the look and feel of images.
\newblock In \emph{Proceedings of the AAAI conference on artificial
  intelligence}, volume~37, pp.\  2555--2563, 2023{\natexlab{b}}.

\bibitem[Wang et~al.(2020)Wang, Li, Zhang, Richardt, and
  Hu]{wang2020transitioning360}
Miao Wang, Yi-Jun Li, Wen-Xuan Zhang, Christian Richardt, and Shi-Min Hu.
\newblock Transitioning360: Content-aware {NFoV} virtual camera paths for
  360{\textdegree} video playback.
\newblock In \emph{2020 IEEE International Symposium on Mixed and Augmented
  Reality (ISMAR)}, pp.\  241--250. IEEE, 2020.

\bibitem[Wang \& Shen(2017)Wang and Shen]{wang2017deepcropping}
Wenguan Wang and Jianbing Shen.
\newblock Deep cropping via attention box prediction and aesthetics assessment.
\newblock In \emph{{IEEE} International Conference on Computer Vision, {ICCV}
  2017, Venice, Italy, October 22-29, 2017}, pp.\  2205--2213. {IEEE} Computer
  Society, 2017.
\newblock \doi{10.1109/ICCV.2017.240}.

\bibitem[Wang et~al.(2019)Wang, Shen, and Ling]{wang2019deepnetwork}
Wenguan Wang, Jianbing Shen, and Haibin Ling.
\newblock A deep network solution for attention and aesthetics aware photo
  cropping.
\newblock \emph{{IEEE} Trans. Pattern Anal. Mach. Intell.}, 41\penalty0
  (7):\penalty0 1531--1544, 2019.
\newblock \doi{10.1109/TPAMI.2018.2840724}.

\bibitem[Wang et~al.(2022)Wang, Wei, Schuurmans, Le, Chi, Narang, Chowdhery,
  and Zhou]{wang2022self}
Xuezhi Wang, Jason Wei, Dale Schuurmans, Quoc Le, Ed~Chi, Sharan Narang,
  Aakanksha Chowdhery, and Denny Zhou.
\newblock Self-consistency improves chain of thought reasoning in language
  models.
\newblock \emph{arXiv preprint arXiv:2203.11171}, 2022.

\bibitem[Wei et~al.(2018)Wei, Zhang, Shen, Lin, M\v{e}ch, Hoai, and
  Samaras]{wei2018goodview}
Zijun Wei, Jianming Zhang, Xiaohui Shen, Zhe Lin, Radom{\'{\i}}r M\v{e}ch, Minh
  Hoai, and Dimitris Samaras.
\newblock Good view hunting: Learning photo composition from dense view pairs.
\newblock In \emph{2018 {IEEE} Conference on Computer Vision and Pattern
  Recognition, {CVPR} 2018, Salt Lake City, UT, USA, June 18-22, 2018}, pp.\
  5437--5446. Computer Vision Foundation / {IEEE} Computer Society, 2018.
\newblock \doi{10.1109/CVPR.2018.00570}.

\bibitem[Wong \& Wong(2025)Wong and Wong]{wong2025aescrop}
Yen{-}Hong Wong and Lai{-}Kuan Wong.
\newblock Aescrop: Aesthetic-driven cropping guided by composition.
\newblock In \emph{{IEEE/CVF} International Conference on Computer Vision,
  {ICCV} 2025 - Workshops, Honolulu, HI, USA, October 19-20, 2025}, pp.\
  3965--3972. {IEEE}, 2025.
\newblock \doi{10.1109/ICCVW69036.2025.00412}.

\bibitem[Wu et~al.(2024)Wu, Zhang, Zhang, Chen, Liao, Li, Gao, Wang, Zhang,
  Sun, Yan, Min, Zhai, and Lin]{wu2024qalign}
Haoning Wu, Zicheng Zhang, Weixia Zhang, Chaofeng Chen, Liang Liao, Chunyi Li,
  Yixuan Gao, Annan Wang, Erli Zhang, Wenxiu Sun, Qiong Yan, Xiongkuo Min,
  Guangtao Zhai, and Weisi Lin.
\newblock Q-align: Teaching lmms for visual scoring via discrete text-defined
  levels.
\newblock In \emph{Forty-first International Conference on Machine Learning,
  {ICML} 2024, Vienna, Austria, July 21-27, 2024}, volume 235 of
  \emph{Proceedings of Machine Learning Research}, pp.\  54015--54029. {PMLR},
  2024.
\newblock URL \url{https://proceedings.mlr.press/v235/wu24ah.html}.

\bibitem[Xiong \& Grauman(2018)Xiong and Grauman]{xiong2018snapangle}
Bo~Xiong and Kristen Grauman.
\newblock Snap angle prediction for 360{\textdegree} panoramas.
\newblock In \emph{Computer Vision - {ECCV} 2018 - 15th European Conference,
  Munich, Germany, September 8-14, 2018, Proceedings, Part {V}}, volume 11209
  of \emph{Lecture Notes in Computer Science}, pp.\  3--20. Springer, 2018.
\newblock \doi{10.1007/978-3-030-01228-1\_1}.

\bibitem[Xiong et~al.(2017)Xiong, Feng, and Zhou]{xiong2017drone}
Xiaoliang Xiong, Jie Feng, and Bingfeng Zhou.
\newblock Automatic view finding for drone photography based on image aesthetic
  evaluation.
\newblock In \emph{Proceedings of the 12th International Joint Conference on
  Computer Vision, Imaging and Computer Graphics Theory and Applications
  {(VISIGRAPP} 2017) - Volume 1: GRAPP, Porto, Portugal, February 27 - March 1,
  2017}, pp.\  282--289. SciTePress, 2017.
\newblock \doi{10.5220/0006255402820289}.

\bibitem[Xu et~al.(2026)Xu, Wei, Zhou, Yin, Deng, Chen, Qu, and
  Cai]{xu2026aes3d}
Chuanzhi Xu, Boyu Wei, Haoxian Zhou, Xuanhua Yin, Zihan Deng, Haodong Chen,
  Qiang Qu, and Weidong Cai.
\newblock Aes3d: Aesthetic assessment in 3d gaussian splatting.
\newblock \emph{arXiv preprint arXiv:2605.05155}, 2026.

\bibitem[Yan et~al.(2013)Yan, Lin, Kang, and Tang]{yan2013change}
Jianzhou Yan, Stephen Lin, Sing~Bing Kang, and Xiaoou Tang.
\newblock Learning the change for automatic image cropping.
\newblock In \emph{2013 {IEEE} Conference on Computer Vision and Pattern
  Recognition, Portland, OR, USA, June 23-28, 2013}, pp.\  971--978. {IEEE}
  Computer Society, 2013.
\newblock \doi{10.1109/CVPR.2013.130}.

\bibitem[Yang et~al.(2023)Yang, Zhou, Cai, Zhang, and Zhang]{yang2023sacnet}
Guo{-}Ye Yang, Wen{-}Yang Zhou, Yun Cai, Song{-}Hai Zhang, and Fang{-}Lue
  Zhang.
\newblock Focusing on your subject: Deep subject-aware image composition
  recommendation networks.
\newblock \emph{Comput. Vis. Media}, 9\penalty0 (1):\penalty0 87--107, 2023.
\newblock \doi{10.1007/s41095-021-0263-3}.

\bibitem[Yang et~al.(2026{\natexlab{a}})Yang, Zhou, Gou, Cui, Li, Gong, Huang,
  and Zhao]{yang2026comex}
Rui Yang, Wei Zhou, Dingyong Gou, Xiaohui Cui, Cong Li, Yinyin Gong, Yipo
  Huang, and Jiliang Zhao.
\newblock {COMEX:} {A} composition-grounded benchmark and learning framework
  for explainable aesthetic image cropping.
\newblock \emph{CoRR}, abs/2608.07570, 2026{\natexlab{a}}.
\newblock \doi{10.48550/ARXIV.2608.07570}.

\bibitem[Yang et~al.(2026{\natexlab{b}})Yang, Wang, Zhang, Xie, Sheng, Chen,
  and Li]{yang2026fine}
Zhichao Yang, Jianjie Wang, Zhixianhe Zhang, Pangu Xie, Xiangfei Sheng, Pengfei
  Chen, and Leida Li.
\newblock Fine-grained image aesthetic assessment: Learning discriminative
  scores from relative ranks.
\newblock In \emph{Proceedings of the IEEE/CVF Conference on Computer Vision
  and Pattern Recognition (CVPR)}, pp.\  145--155, June 2026{\natexlab{b}}.

\bibitem[Yao et~al.(2025)Yao, Zheng, Yuan, Cai, Wu, Chen, Li, and
  Jiang]{yao2025ppc}
Lujian Yao, Siming Zheng, Xinbin Yuan, Zhuoxuan Cai, Pu~Wu, Jinwei Chen, Bo~Li,
  and Peng-Tao Jiang.
\newblock Photography perspective composition: Towards aesthetic perspective
  recommendation.
\newblock In \emph{Advances in Neural Information Processing Systems},
  volume~38, 2025.
\newblock \doi{10.52202/085713-3020}.
\newblock URL
  \url{https://proceedings.neurips.cc/paper_files/paper/2025/hash/8227285e32f70e07fa3a247f3a48006d-Abstract-Conference.html}.

\bibitem[You et~al.(2026)You, Wang, Zhang, Cai, Gu, Xue, Dong, and
  Zhang]{you2026photoframer}
Zhiyuan You, Ke~Wang, He~Zhang, Xin Cai, Jinjin Gu, Tianfan Xue, Chao Dong, and
  Zhoutong Zhang.
\newblock {PhotoFramer}: Multi-modal image composition instruction.
\newblock In \emph{Proceedings of the IEEE/CVF Conference on Computer Vision
  and Pattern Recognition}, pp.\  10197--10207, 2026.

\bibitem[Yu et~al.(2018)Yu, Lee, Na, Kang, and Kim]{yu2018deep}
Youngjae Yu, Sangho Lee, Joonil Na, Jaeyun Kang, and Gunhee Kim.
\newblock A deep ranking model for spatio-temporal highlight detection from a
  360{\textdegree} video.
\newblock In \emph{Proceedings of the AAAI Conference on Artificial
  Intelligence}, volume~32, 2018.

\bibitem[Yuan et~al.(2024)Yuan, Li, and Chen]{yuan2024vlpcrop}
Quan Yuan, Leida Li, and Pengfei Chen.
\newblock Aesthetic image cropping meets {VLP:} enhancing good while reducing
  bad.
\newblock \emph{J. Vis. Commun. Image Represent.}, 105:\penalty0 104316, 2024.
\newblock \doi{10.1016/J.JVCIR.2024.104316}.

\bibitem[Yuan et~al.(2026)Yuan, Huang, Chen, and Li]{yuan2026humancrop}
Quan Yuan, Yipo Huang, Pengfei Chen, and Leida Li.
\newblock Humancrop-thinker: An inference-driven framework with explicit
  thinking for explainable human-centric image cropping.
\newblock \emph{Expert Syst. Appl.}, 316:\penalty0 131664, 2026.
\newblock \doi{10.1016/J.ESWA.2026.131664}.

\bibitem[Zadrozny \& Elkan(2002)Zadrozny and Elkan]{zadrozny2002transforming}
Bianca Zadrozny and Charles Elkan.
\newblock Transforming classifier scores into accurate multiclass probability
  estimates.
\newblock In \emph{Proceedings of the eighth ACM SIGKDD international
  conference on Knowledge discovery and data mining}, pp.\  694--699, 2002.

\bibitem[Zelikman et~al.(2022)Zelikman, Wu, Mu, and Goodman]{zelikman2022star}
Eric Zelikman, Yuhuai Wu, Jesse Mu, and Noah Goodman.
\newblock {STaR}: Bootstrapping reasoning with reasoning.
\newblock \emph{Advances in Neural Information Processing Systems},
  35:\penalty0 15476--15488, 2022.

\bibitem[Zeng et~al.(2019)Zeng, Li, Cao, and Zhang]{zeng2019grid}
Hui Zeng, Lida Li, Zisheng Cao, and Lei Zhang.
\newblock Reliable and efficient image cropping: {A} grid anchor based
  approach.
\newblock In \emph{{IEEE} Conference on Computer Vision and Pattern
  Recognition, {CVPR} 2019, Long Beach, CA, USA, June 16-20, 2019}, pp.\
  5949--5957. Computer Vision Foundation / {IEEE}, 2019.
\newblock \doi{10.1109/CVPR.2019.00610}.

\bibitem[Zeng et~al.(2022)Zeng, Li, Cao, and Zhang]{zeng2022gaicd}
Hui Zeng, Lida Li, Zisheng Cao, and Lei Zhang.
\newblock Grid anchor based image cropping: {A} new benchmark and an efficient
  model.
\newblock \emph{{IEEE} Trans. Pattern Anal. Mach. Intell.}, 44\penalty0
  (3):\penalty0 1304--1319, 2022.
\newblock \doi{10.1109/TPAMI.2020.3024207}.

\bibitem[Zhang et~al.(2021{\natexlab{a}})Zhang, Niu, and Zhang]{zhang2021image}
Bo~Zhang, Li~Niu, and Liqing Zhang.
\newblock Image composition assessment with saliency-augmented multi-pattern
  pooling.
\newblock \emph{arXiv preprint arXiv:2104.03133}, 2021{\natexlab{a}}.

\bibitem[Zhang et~al.(2022)Zhang, Niu, Zhao, and Zhang]{zhang2022humancentric}
Bo~Zhang, Li~Niu, Xing Zhao, and Liqing Zhang.
\newblock Human-centric image cropping with partition-aware and
  content-preserving features.
\newblock In \emph{Computer Vision - {ECCV} 2022 - 17th European Conference,
  Tel Aviv, Israel, October 23-27, 2022, Proceedings, Part {VII}}, volume 13667
  of \emph{Lecture Notes in Computer Science}, pp.\  181--197. Springer, 2022.
\newblock \doi{10.1007/978-3-031-20071-7\_11}.

\bibitem[Zhang et~al.(2026{\natexlab{a}})Zhang, Ding, Jiang, Chen, Zharkov,
  Patel, and Liang]{zhang2026procrop}
Ke~Zhang, Tianyu Ding, Jiachen Jiang, Tianyi Chen, Ilya Zharkov, Vishal~M.
  Patel, and Luming Liang.
\newblock Procrop: Learning aesthetic image cropping from professional
  compositions.
\newblock In \emph{Fortieth {AAAI} Conference on Artificial Intelligence,
  {AAAI} 2026, Singapore, January 20-27, 2026}, pp.\  12600--12608. {AAAI}
  Press, 2026{\natexlab{a}}.
\newblock \doi{10.1609/aaai.v40i15.38255}.

\bibitem[Zhang et~al.(2021{\natexlab{b}})Zhang, Qin, Wang, and
  Li]{zhang2021rethinking}
Yi~Zhang, Hongwei Qin, Xiaogang Wang, and Hongsheng Li.
\newblock Rethinking noise synthesis and modeling in raw denoising.
\newblock In \emph{Proceedings of the IEEE/CVF International Conference on
  Computer Vision (ICCV)}, pp.\  4593--4601, October 2021{\natexlab{b}}.
\newblock \doi{10.1109/ICCV48922.2021.00455}.

\bibitem[Zhang et~al.(2018)Zhang, Xu, Yu, and Gao]{zhang2018saliency}
Zhiheng Zhang, Yanyu Xu, Jingyi Yu, and Shenghua Gao.
\newblock Saliency detection in 360{\textdegree} videos.
\newblock In \emph{European conference on computer vision}, pp.\  504--520.
  Springer, 2018.

\bibitem[Zhang et~al.(2026{\natexlab{b}})Zhang, Liu, Liu, Li, Zhang, Kong, and
  Zhang]{zhang2026discrete}
Ziqing Zhang, Xiao Liu, Kai Liu, Jianze Li, Weihang Zhang, Linghe Kong, and
  Yulun Zhang.
\newblock Discrete annotation, continuous preference: Rethinking supervision
  for accurate and generalizable aesthetic image cropping.
\newblock \emph{arXiv preprint arXiv:2610.00582}, 2026{\natexlab{b}}.

\bibitem[Zhao et~al.(2025)Zhao, Lu, Zhang, Li, Li, Liu, Hu, Chen, Wang, and
  Guo]{zhao2025composition}
Zhaoran Zhao, Peng Lu, Anran Zhang, Peipei Li, Xia Li, Xuannan Liu, Yang Hu,
  Shiyi Chen, Liwei Wang, and Wenhao Guo.
\newblock Can machines understand composition? dataset and benchmark for
  photographic image composition embedding and understanding.
\newblock In \emph{2025 {IEEE/CVF} Conference on Computer Vision and Pattern
  Recognition, {CVPR} 2025}, pp.\  14411--14421. Computer Vision Foundation /
  {IEEE}, 2025.

\bibitem[Zhong et~al.(2023)Zhong, Cheng, Wu, Yuan, Zheng, Li, Hu, Lin, Sato,
  and Sato]{zhong2023clipcrop}
Zhihang Zhong, Mingxi Cheng, Zhirong Wu, Yuhui Yuan, Yinqiang Zheng, Ji~Li, Han
  Hu, Stephen Lin, Yoichi Sato, and Imari Sato.
\newblock Clipcrop: Conditioned cropping driven by vision-language model.
\newblock In \emph{{IEEE/CVF} International Conference on Computer Vision,
  {ICCV} 2023 - Workshops, Paris, France, October 2-6, 2023}, pp.\  294--304.
  {IEEE}, 2023.
\newblock \doi{10.1109/ICCVW60793.2023.00037}.

\bibitem[Zhou et~al.(2024)Zhou, Wang, Lin, Su, Chen, Tao, Zheng, Yuan, Wan, and
  Zhang]{zhou2024uniaa}
Zhaokun Zhou, Qiulin Wang, Bin Lin, Yiwei Su, Rui Chen, Xin Tao, Amin Zheng,
  Li~Yuan, Pengfei Wan, and Di~Zhang.
\newblock {UNIAA:} {A} unified multi-modal image aesthetic assessment baseline
  and benchmark.
\newblock \emph{CoRR}, abs/2404.09619, 2024.
\newblock \doi{10.48550/ARXIV.2404.09619}.

\bibitem[Zhu et~al.(2025)Zhu, Yan, Zhou, Yao, Shao, Zhao, and
  Li]{zhu2025attribute}
Hancheng Zhu, Li~Yan, Yong Zhou, Rui Yao, Zhiwen Shao, Jiaqi Zhao, and Leida
  Li.
\newblock Image cropping with content and composition attribute-aware global
  relation reasoning.
\newblock \emph{{ACM} Trans. Multim. Comput. Commun. Appl.}, 21\penalty0
  (4):\penalty0 125:1--125:19, 2025.
\newblock \doi{10.1145/3719012}.

\end{thebibliography}
\bibliographystyle{iclr2027_conference}

\appendix
\section{Appendix}

This document provides supplementary material for the paper “PhotoGuider: Composition-Aware Cropping and 360\textdegree{}~View Framing with Explainable Photographic Assessments” to complement the main manuscript. Sec. \ref{a1}, \ref{a2}, \ref{a3}, and \ref{a4} provide details of the photographic tokens, refinement gates, MLLM consultant, and Explainer, respectively. Sec. \ref{a5} offers implementation details of PhotoGuider. Sec. \ref{a6} provides the ablation study of modules in the refinement stage. Sec. \ref{a7} gives the detailed analysis of user preferences for 360\textdegree{}~framings. 

\paragraph{Supplementary material.} The project page (\url{https://ntuneillee.github.io/research/photoguider/}) provides two videos of the full 360\ensuremath{^{\circ}} pipeline, from
direction search to the final native re-render with its explanation, and the ten user-study
scenes in which our framing was preferred most strongly, each showing the panorama, the three
methods' framings with their first-place shares and camera parameters, our four token
assessments, and the three direction fields. All scenes are CC0 panoramas from Poly Haven.

\subsection{The four photographic tokens}\label{a1}

\begin{wrapfigure}{r}{0.48\textwidth}
    \vspace{-0.8\baselineskip}
    \centering
    \includegraphics[width=\linewidth]{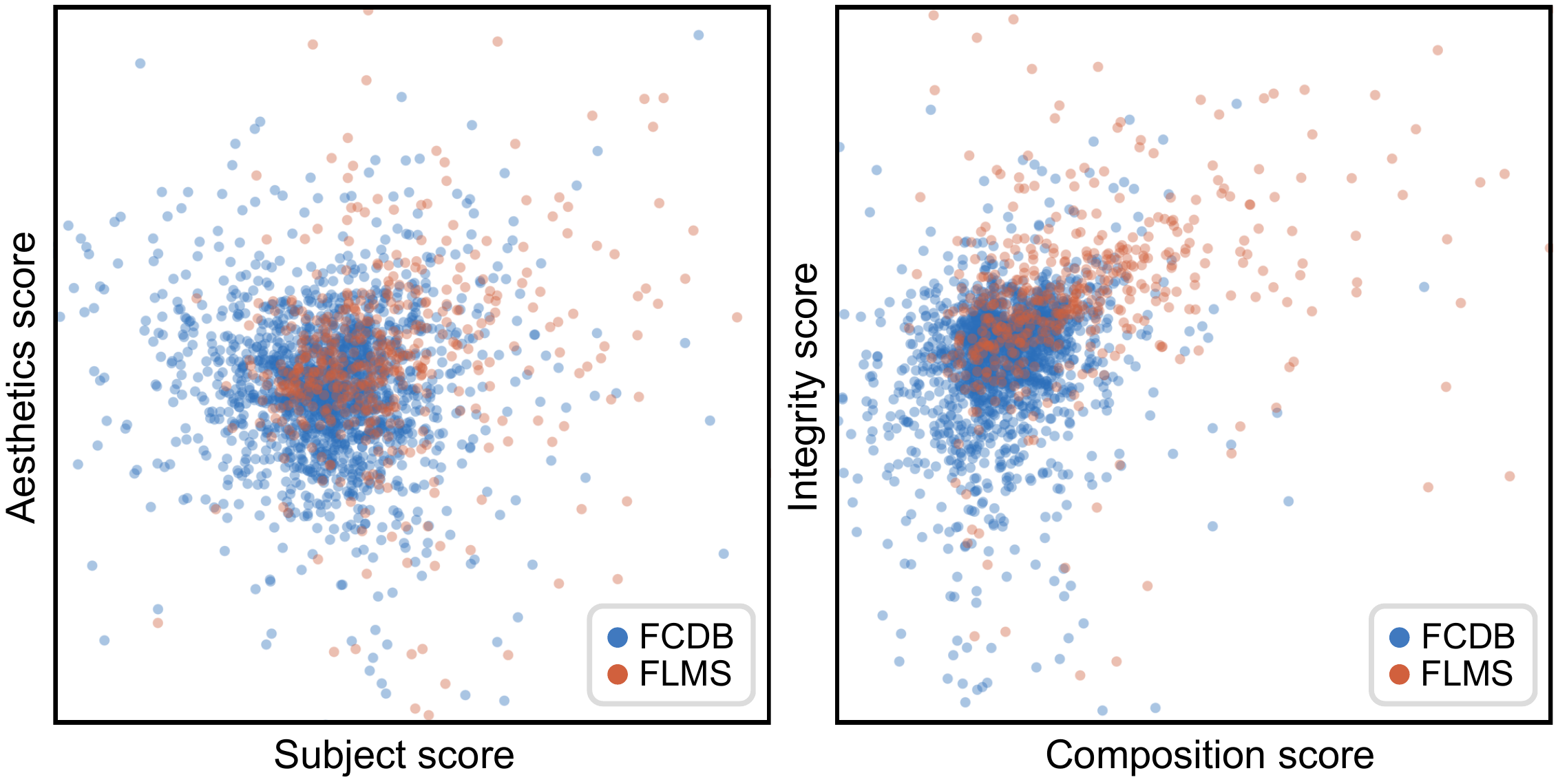}
    
    \vspace{-3.5pt}
    \caption{\textbf{Dataset Bias from FCDB and FLMS datasets.} Each point is one crop selected by PhotoGuider, positioned by the contributions of two of the four photographic tokens. Normalized by the pooled standard deviation, the median-to-median displacement is 0.34 SD in the S-A plane and 0.47 SD in the C-I plane.}
    \label{fig:datasetbias}
    \vspace{-0.5\baselineskip}
\end{wrapfigure}

\textbf{Shared photographic features.} We measure 60 features on each perspective frame and group them into four photographic tokens: Subject (S), Composition (C), Aesthetics (A), and Integrity (I). The same feature groups support direction selection, crop ranking, and explanation. Figure~\ref{fig:datasetbias} shows differences in token contributions between crops selected on FCDB and FLMS. We evaluate transfer across these benchmarks using the same trained checkpoint (Sec.~\ref{sec:mainresults}). \WFclear

\textbf{Subject: what is being photographed.} A photographer first checks whether the subject is fully in frame and how much of the frame it
occupies. We measure this from three complementary views. The self-supervised attention of
{DINOv3} supplies saliency coverage and peak (\texttt{vmix\_\allowbreak{}max}, \texttt{v1\_\allowbreak{}max}, \texttt{v1\_\allowbreak{}cov}). {InSPyReNet}
supplies the subject's precise outline and completeness (\texttt{inspy\_\allowbreak{}capture}, \texttt{inspy\_\allowbreak{}max}). The
{SAM} object mask corresponding to the subject supplies its area share, centering, and
agreement with the InSPyReNet reading (\texttt{sam\_\allowbreak{}mask\_\allowbreak{}area\_\allowbreak{}ratio}, \texttt{sam\_\allowbreak{}mask\_\allowbreak{}centered\_\allowbreak{}x/y},
\texttt{sam\_\allowbreak{}mask\_\allowbreak{}vs\_\allowbreak{}inspy\_\allowbreak{}iou}), the last being a measure of how far two subject
readings agree. A geometric term completes the group with subject size (\texttt{subject\_\allowbreak{}size\_\allowbreak{}ratio}).

\textbf{Composition: how the subject is placed.} With the subject in frame, a photographer checks placement against a textbook vocabulary: the
rule of thirds, symmetry, triangular structure, balance of visual weight, negative space and
shot scale. Three signals are complementary. Geometry encodes the rules directly (\texttt{thirds\_\allowbreak{}align}, \texttt{sym\_\allowbreak{}v\_\allowbreak{}diff}, \texttt{sym\_\allowbreak{}diag\_\allowbreak{}diff}, \texttt{tri\_\allowbreak{}top\_\allowbreak{}apex}, \texttt{tri\_\allowbreak{}bot\_\allowbreak{}apex},
\texttt{visual\_\allowbreak{}weight\_\allowbreak{}x/y}, \texttt{ca\_\allowbreak{}dist}, \texttt{area\_\allowbreak{}ratio}, \texttt{log\_\allowbreak{}aspect\_\allowbreak{}diff}). {SAM thirds alignment} measures how segmented objects align with the rule-of-thirds structure (\texttt{sam\_\allowbreak{}thirds\_\allowbreak{}q\_\allowbreak{}mean/max}, \texttt{sam\_\allowbreak{}thirds\_\allowbreak{}vs\_\allowbreak{}center}). {CLIP} multi-prompt similarity
(\texttt{clip\_\allowbreak{}sim\_\allowbreak{}\{pro, balanced, leading, thirds, symmetric, negspace, centered, dynamic\}},
\texttt{clip\_\allowbreak{}max\_\allowbreak{}sim}) provides a computable linguistic anchor for the vocabulary itself: eight prompts
are eight rules stated in natural language, and CLIP's image-text alignment turns
"does this look like a leading-lines composition" into a measurable number.

\textbf{Integrity: what the frame cut off.} Last comes the reflex check of a professional: the crop must not sever a hand or a head, must
not leave a mutilated object, and must keep the depth layering clean. This check evaluates object boundaries throughout the frame, independently of subject identity. That is SAM's most direct use, and all 20 dimensions of this token come
from it. Encoder-side embedding statistics (\texttt{sam\_\allowbreak{}emb\_\allowbreak{}norm/var/kurt},
\texttt{sam\_\allowbreak{}cand\_\allowbreak{}norm\_\allowbreak{}mean/max/std}, \texttt{sam\_\allowbreak{}cand\_\allowbreak{}to\_\allowbreak{}outside\_\allowbreak{}ratio}, \texttt{sam\_\allowbreak{}image\_\allowbreak{}total\_\allowbreak{}norm}) characterize
how much information the crop holds and how it contrasts with what lies outside;
\texttt{sam\_\allowbreak{}cos\_\allowbreak{}to\_\allowbreak{}image} is a continuous measure of shot scale, from close-up to full view. On the
decoder side we count the objects (\texttt{sam\_\allowbreak{}obj\_\allowbreak{}count\_\allowbreak{}in\_\allowbreak{}cand}, \texttt{sam\_\allowbreak{}obj\_\allowbreak{}top1\_\allowbreak{}share},
\texttt{sam\_\allowbreak{}obj\_\allowbreak{}top3\_\allowbreak{}share}, \texttt{sam\_\allowbreak{}obj\_\allowbreak{}area\_\allowbreak{}cv}, \texttt{sam\_\allowbreak{}dominant\_\allowbreak{}ratio}), check whether each remains
cleanly definable (\texttt{sam\_\allowbreak{}mask\_\allowbreak{}q\_\allowbreak{}mean/std/margin}, whose quality and boundary margin drop when a
cropping line severs an object), and read the scene's depth layering (\texttt{sam\_\allowbreak{}h\_\allowbreak{}area\_\allowbreak{}cv},
\texttt{sam\_\allowbreak{}h\_\allowbreak{}iou\_\allowbreak{}top12}, \texttt{sam\_\allowbreak{}h\_\allowbreak{}cand\_\allowbreak{}match\_\allowbreak{}max}).

\textbf{Aesthetics: whether it is pleasing.} The Aesthetics token measures visual appeal using two complementary aesthetic priors.
The {LAION aesthetic head} (\texttt{laion\_\allowbreak{}score}, \texttt{laion\_\allowbreak{}pi\_\allowbreak{}minmax}) represents a crowd's sense that
an image is appealing, and {FGAesQ} (\texttt{fgaesq\_\allowbreak{}score}) represents a fine-grained expert judgment
of where the appeal lies; colour saturation (\texttt{color\_\allowbreak{}sat\_\allowbreak{}mean}) supplements them. The last group
measures the subject-to-background separation in colour and tone that SAM's parse makes
computable (\texttt{sam\_\allowbreak{}fg\_\allowbreak{}bg\_\allowbreak{}sat\_\allowbreak{}diff}, \texttt{sam\_\allowbreak{}fg\_\allowbreak{}bg\_\allowbreak{}val\_\allowbreak{}diff}, \texttt{sam\_\allowbreak{}fg\_\allowbreak{}bg\_\allowbreak{}lab\_\allowbreak{}dist}, \texttt{sam\_\allowbreak{}fg\_\allowbreak{}var}).
What a photographer calls the subject "popping out" is, concretely, a contrast in saturation,
value and Lab colour between a foreground object and its background, and computing it requires
that object to have been separated first.

\textbf{Token Assessments.} The ranker operates as a gradient-boosted tree model explained post hoc. For a
given frame, TreeSHAP~\cite{lundberg2020treeshap} decomposes each model's raw output into
feature attributions. These attributions pass through the same per-image normalization and
model-output weighting as the final score, giving the approximate decomposition

\begin{equation}
q\approx\sum_{g\in\{\mathrm{S},\mathrm{C},\mathrm{A},\mathrm{I}\}}\Phi_g+\mathrm{const},
\qquad
\Phi_g=\sum_{j\in\mathcal{F}_g}\phi_j ,
\end{equation}

where $\phi_j$ is the weighted attribution of feature $j$ and $\mathcal{F}_g$ is the disjoint
index set of token $g$. The attribution decomposition is exact for three uncalibrated heads carrying a combined weight of 0.88. For the calibrated head, which carries the remaining weight of 0.12, attributions are propagated through the isotonic mapping using a secant approximation. Token attributions aggregate the feature-level shares of cross-token interactions, subject to this approximation. Finally, each $\Phi_g$ is mapped to a
percentile rank against the training reference distribution and reported independently,
without any additive assumption.

\textbf{Candidate Generation.} With a target ratio, about 300 candidates come from eight sources: a base geometric grid
re-aspected to the ratio; anchors on the subject mask and on the top saliency peaks, each
crossed with seven medium areas; a multi-scale pass over the best-covering candidates; and
four composition-principle generators (diagonal, negative space, leading line, boundary).
Without a ratio, a grid of seven scales and 36 positions is laid over each of eight ratios,
the seven standard ratios 16:9, 3:2, 4:3, 1:1, 3:4, 2:3 and 9:16 plus the source image's
own ratio, deduplicated within 2\%. The eight sources are described in Table.\ref{tab:generation}.

\newsavebox{\moduleAblationBox}
\begin{table}[t]
\centering\small
\caption{The eight candidate sources of the aspect ratio-aware pool. Every source is deduplicated by non-maximum suppression at IoU 0.95.}\label{tab:generation}

\vspace{5pt}

\begin{tabularx}{\textwidth}{@{}>{\raggedright\arraybackslash}p{0.20\textwidth} >{\raggedright\arraybackslash}p{0.25\textwidth} >{\raggedright\arraybackslash}X@{}}
\toprule
Source & Anchor & Construction \\
\midrule
Base grid & image geometry & center-shift grid, small-area grid and seeded random boxes, each re-aspected to the target ratio \\
Subject anchors & InSPyReNet subject mask & mask centroid, blobs and bounding box $\times$ seven medium areas \\
Saliency peaks & DINOv3 saliency map & top-five peaks $\times$ seven medium areas \\
Multi-scale & the three sources above & top-20 by saliency coverage $\times$ scale factors $\{0.7, 0.85, 1.15, 1.3\}$ \\
Diagonal & subject mask and saliency & frames placing the subject on an image diagonal \\
Negative space & subject mask and saliency & frames leaving open space opposite the subject \\
Leading line & saliency structure & frames aligned with a dominant linear structure \\
Boundary & subject mask & frames set by the subject's extent with a margin \\
\bottomrule
\end{tabularx}

\label{tab:pool_sources}
\end{table}

\subsection{Gated refinement} \label{a2}

The refinement stage applies gated spatial adjustments to the selected frame $c^{*}$ to obtain $b$. Each adjustment is evaluated by a LightGBM model that predicts its IoU gain on the current image. The adjustment is applied when the predicted gain exceeds the corresponding threshold.

\textbf{Recenter.} The gate chooses where to place the frame center. Two candidate targets
are available: $t_{\mathrm{thirds}}$, the saliency centroid softly aligned to the nearest
rule-of-thirds point, and $t_{\mathrm{cam}}$, the centroid of CLIP's compositional attention.
The gate predicts $\hat m$, the IoU advantage of the CLIP target over the thirds target, and
turns it into a mixing weight $w=\mathrm{clip}(\hat m/T_{g},\,0,\,1)$, where $T_{g}$ is a fixed
scale constant. The target is the mix $t=(1-w)\,t_{\mathrm{thirds}}+w\,t_{\mathrm{cam}}$, and
the center $\mathbf{x}$ of the frame moves a fixed fraction $\beta$ of the way toward it.

\textbf{Expand.} Four independent gates $k\in\{\text{cov},\text{incl},\Delta\text{aes},
\Delta\text{gold}\}$ each predict the IoU gain $\hat g_{k}(s)$ of enlarging the frame by a
factor $s\in\mathcal{S}$ about its centre, at fixed aspect ratio. They read four different
signals: the coverage gate reads how much of the subject's saliency the current frame
captures; the inclination gate reads CLIP's relative response to wide versus tight
composition prompts; the $\Delta$-aesthetics gate reads the change of the LAION and FGAesQ
scores between the current frame and the enlarged one; the $\Delta$-golden gate reads the
change in the subject's distance to the nearest golden-section point. A gate fires when
$\max_{s}\hat g_{k}(s)>\tau_{e}$, and the frame is enlarged to the factor with the largest
predicted gain among the gates that fired. The two $\Delta$ gates use score changes across scales. In our measurements, static scores provide little signal for expansion. We hypothesize that using score changes contributes to cross-dataset transfer. Each of the four gates evaluates a distinct signal and applies its own expansion criterion.

\textbf{Shrink.} When no expand gate fires, one fallback gate reads the shot-scale measure (the cosine between the frame's SAM embedding and the whole image's) and, above its own
threshold $\tau_{s}$, tightens by a factor $s\in\mathcal{S}_{\mathrm{shr}}$ a frame that looks
too much like the whole image.

\subsection{MLLM as Consultant}\label{a3}

In the final stage, we incorporate Venus~\cite{du2026venus} as an MLLM consultant to propose candidate boxes, restricting it from directly scoring or making final decisions in our pipeline.

\paragraph{Self-Consistency.} A primary prompt is decoded $1 + m$ times (one greedy decoding alongside $m$ temperature-sampled outputs with fixed seeds), where the coordinate-wise median yields the core opinion $\tilde{v}_{\mathrm{c}}$. To build a broader opinion $\tilde{v}_{\mathrm{f}}$, we compute the median across the union of responses from $M$ paraphrased prompts (decoded greedily with whole-image predictions filtered out). The coordinate-wise median aggregates the decoded box predictions.

\paragraph{Twin Candidates.} For each shortlist candidate $c_i$ that aligns with the consultant, defined by $\mathrm{IoU}(c_i, \tilde{v}_{\mathrm{c}}) \ge 0.5$, we construct a twin candidate $c_i^{t}$. Its center and area are shifted toward $\tilde{v}_{\mathrm{c}}$ within fixed bounds at a constant ratio, followed by a complete recomputation of all 60 features. The frozen ranker then evaluates both the original and twin candidates on the merged pool, switching to the twin only when a substantial margin is achieved.

\paragraph{Bounded Concession.} Prior to final delivery, if the refined box $b$ agrees with the broader opinion ($\mathrm{IoU}(b, \tilde{v}_{\mathrm{f}}) \ge \tau$), the system interpolates the center linearly and the area geometrically:
\begin{equation}
\mathbf{x}_{\hat{b}} = (1 - \lambda_c)\mathbf{x}_b + \lambda_c\mathbf{x}_{\tilde{v}_{\mathrm{f}}}, \quad A_{\hat{b}} = A_b^{1 - \lambda_a} A_{\tilde{v}_{\mathrm{f}}}^{\lambda_a}.
\end{equation}
The log aspect ratio is adjusted via $\lambda_r$: setting $\lambda_r = 0$ when a fixed target ratio is specified (preserving the output ratio), or $\lambda_r = 1$ when unconstrained, thereby adopting the consultant's aspect ratio opinion to refine the discrete enumeration from the main paper's ``Candidate generation and shortlisting'' section. Below the threshold $\tau$, the consultant is considered to be proposing a distinct photographic composition, and no adjustments are adopted. All hyper-parameters are fitted on the training split under a bounded-regression constraint without per-image learned decisions.

\subsection{Explanation}\label{a4}

\paragraph{Measurements as Shared Foundation.} Our explainer cites the quantitative features used by the ranker. Token assessments and feature attribution lists originate from TreeSHAP~\cite{lundberg2017shap,lundberg2020treeshap} decomposition of the ranker's output, while geometric assertions are computed from the same underlying vision features.

\paragraph{Training via Distillation Filtering.} We adopt Qwen2.5-VL-7B~\cite{bai2025qwen25vl} as the explainer MLLM, using the same base model as a teacher to draft explanations for training images. To build a high-quality dataset, drafts must pass four distillation filters: a \emph{Silence Filter} discarding geometric claims without valid subject masks, a \emph{Consistency Filter} rejecting values or directions conflicting with measurements, an \emph{Evidence Filter} enforcing minimum feature citations beyond raw geometry, and a \emph{Wording Filter} eliminating banned phrasing or incorrect languages. The student model learns exclusively from passed drafts via QLoRA~\cite{dettmers2023qlora,hu2022lora}.

\paragraph{Inference Prompt Construction.} At inference, the explainer receives the input image, the final cropped frame, and a structured prompt comprising the four token assessments, frame geometry, and key feature attributions mapped to human-interpretable photographic terms. Anchor sentences derived directly from physical measurements are appended to the prompt to constrain the corresponding statements; they do not guarantee the correctness of the entire generated explanation.

\paragraph{Contradiction rate.} We measure the consistency between generated statements and the supplied measurements. We report the contradiction rate detected by a rule-based checker. Anchor sentences constrain the corresponding statements, while the remaining sentences are generated freely. Appendix~\ref{a7} reports a complementary human preference study on composition explanations.

\begin{table}[t]
\centering\small

\caption{Module-level ablation.}
\label{tab:module_ablation}

\vspace{5pt}

\global\setbox\moduleAblationBox=\hbox{
\begin{tabular}{l | cc cc}
\toprule
& \multicolumn{2}{c}{FCDB} & \multicolumn{2}{c}{FLMS} \\
\cmidrule(lr){2-3}
\cmidrule(lr){4-5}
Configuration & BDE$\downarrow$ & mIoU$\uparrow$ & BDE$\downarrow$ & mIoU$\uparrow$ \\
\midrule

\multicolumn{5}{l}{\emph{Refinement variants}} \\

No refinement                     & 0.060 & 0.754 & 0.044 & 0.820 \\
Recenter only                     & 0.059 & 0.758 & 0.043 & 0.824 \\
No recenter                       & 0.060 & 0.755 & 0.042 & 0.832 \\
Ungated expansion                 & 0.066 & 0.738 & 0.042 & 0.834 \\
Recenter target fixed to thirds   & 0.059 & 0.758 & \textbf{0.041} & 0.834 \\
Recenter target fixed to CLIP     & 0.060 & 0.750 & 0.042 & 0.832 \\
PhotoGuider (full refinement) \ \ \ \        & \textbf{0.058} & \textbf{0.759} & \textbf{0.041} & \textbf{0.835} \\

\bottomrule
\end{tabular}
}
\usebox{\moduleAblationBox}
\end{table}

\subsection{Implementation Details}\label{a5}

\paragraph{View scorer.} The view scorer is implemented as a hierarchical additive network operating over the 60 standardized features. Each feature is processed by an individual subnetwork ($1 \to 32 \to 16 \to 1$ with ELU activation), supplemented by pair subnetworks ($2 \to 16 \to 8 \to 1$) for 20 selected feature pairs. To ensure exact score additive decomposition, each feature output is scaled by a learned token weight, with pairwise terms split equally between the tokens of their constituent features. The architecture contains $4.0 \times 10^4$ parameters and is trained on the GAICD training set using a within-image pairwise ranking loss, applied to crop pairs from the same image with MOS differences of at least 0.1. Optimization employs Adam (learning rate $10^{-3}$, weight decay $10^{-5}$, pair batch size 8,192) with early stopping based on validation SRCC (patience 15), where the optimal checkpoint is selected at epoch 60.

\textbf{Ranker.} Training uses a learning rate of 0.05. The model combines three component scores $q_A$ (0.6 regression / 0.4 ranking), $q_B$ (0.55 / 0.45), and $q_C$ (0.6 / 0.4) with isotonic calibration into a final score $q = 0.4 q_A + 0.4 q_B + 0.2 q_C$, totaling approximately $2.1 \times 10^4$ parameters. 

\textbf{Refinement.} Recentering operates with $\beta = 0.20$, rule-of-thirds weight 0.4, and gate margin $T_g = 0.05$. Four expansion gates and a single shrink gate are LightGBM regressors trained on IoU gain across scale candidate sets $S = \{1.1, 1.3, 1.5, 2.0\}$ and $\{0.6, 0.7, 0.8, 0.9\}$ ($\approx 4.2 \times 10^4$ parameters). Refinement thresholds $(\tau_e, \tau_s)$ are set to $(0.03, 0.05)$ given a target aspect ratio, and $(-0.02, 0.08)$ otherwise. The complete decision layer contains roughly $6.3 \times 10^4$ parameters, executing within milliseconds on a CPU. 

Note that the free-form expansion threshold is $-0.02$: the gate's gain estimates are conservative on free-form pools, and the training-split grid search selected a threshold that admits
expansions with slightly negative predicted gain.

\textbf{Consultant.} Venus is quantized in 4-bit NF4. We run $1 + m$ decodes ($m = 5$, top-$p$ 0.9, temperature 1.0, fixed seeds) alongside $M = 4$ paraphrases, producing 10 decodes per image. Twin candidates require $\mathrm{IoU} \ge 0.5$. Final blending parameters $(\lambda_c, \lambda_a, \tau)$ are set to $(0.4, 0.2, 0.5)$ with a target ratio, and $(0.7, 0.3, 0.45)$ without (adopting the consultant aspect ratio). All hyperparameters are determined via grid search on the training split without consulting FCDB or FLMS test sets. 

\textbf{Explainer.} We fine-tune Qwen2.5-VL-7B-Instruct using QLoRA ($r = 16$, $\alpha = 32$, dropout 0.05) applied to language-model linear layers while freezing the vision tower and aligner, yielding 40.4M trainable parameters (0.8\% of total). Training uses NF4 base with bf16 compute, learning rate $2 \times 10^{-4}$ with cosine decay, gradient accumulation step 32, and sequence length 3,072. 

\subsection{Module Ablation}\label{a6}

Table~\ref{tab:module_ablation} isolates the performance contribution of each refinement stage. Compared with no refinement, full refinement improves mIoU from 0.754 to 0.759 on FCDB and from 0.820 to 0.835 on FLMS, while reducing BDE on both benchmarks.

\begin{table}[t]
\centering
\small
\renewcommand{\arraystretch}{1.1}
\setlength{\abovecaptionskip}{2pt}
\caption{{Choices per system in the blind three-way study.}}

\vspace{5pt}

\setlength{\tabcolsep}{3pt}
\begin{tabularx}{\wd\moduleAblationBox}{@{}l*{3}{>{\centering\arraybackslash}X}@{}}
    \toprule
    Image & PhotoGuider & Venus & ArtiMuse \\
    \midrule
    \texttt{fcdb\_16002047337} & \textbf{20} & 16 & 14 \\
    \texttt{fcdb\_9349007347}  & \textbf{20} & 15 & 15 \\
    \texttt{flms\_15659590}    & \textbf{20} & 19 & 11 \\
    \texttt{flms\_799562644}   & \textbf{21} & 15 & 14 \\
    \texttt{flms\_1588688990}  & \textbf{19} & 17 & 14 \\
    \midrule
    All (250) & \textbf{100} (40.0\%) & 82 (32.8\%) & 68 (27.2\%) \\

\bottomrule
\end{tabularx}
\label{tab:explanation_user_study}
\end{table}

\begin{figure}[t]
  \centering
  \includegraphics[width=\linewidth]{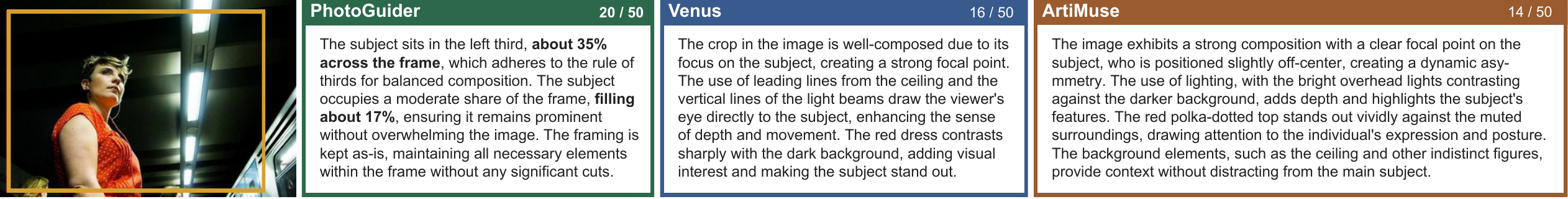}
  \vspace{-18pt}
  \caption{\textbf{An example from the human preference study on composition explanations.} PhotoGuider, Venus, and ArtiMuse explain the same crop. The counts above each explanation indicate how many of the 50 participants selected it as best explaining the crop's compositional aesthetics. Method names are shown for reference and were concealed during the study.}
  \label{fig:explanation_demo}
\end{figure}

\subsection{Analysis of user study}\label{a7}

\textbf{Composition Explanations.} Five images were randomly selected:
\texttt{fcdb\_16002047337}, \texttt{fcdb\_9349007347}, \texttt{flms\_15659590}, \texttt{flms\_799562644}, and \texttt{flms\_1588688990}.
For each image, all three systems explain the same PhotoGuider crop, holding the visual input fixed across the three explanations. The study followed a double-blind design. Table~\ref{tab:explanation_user_study} gives the overall and per-image counts.
PhotoGuider's explanation received 100 of 250 choices (40.0\%), Venus 82 (32.8\%), and ArtiMuse 68 (27.2\%). PhotoGuider's observed choice share exceeded those of Venus and ArtiMuse by 7.2 and 12.8 percentage points, respectively. 

The result supports that: when users judge how well a text explains the composition of a given crop, an explanation whose content is anchored in measured compositional quantities is chosen at least as often as the explanation of a dedicated aesthetic MLLM and more often than ArtiMuse's. The texts themselves show why the preference splits. ArtiMuse frequently targets irrelevant visual elements rather than composition itself. An obvious example is found in Image 1 (Fig. \ref{fig:explanation_demo}), where it describes the plain red top as “polka-dotted” and mistakenly cites this as justification for a favorable compositional score. Venus stays on composition but in a template: four of five passages open with “The crop in the image is well-composed due to its focus on”, all are wholly positive, and none gives a position or size. PhotoGuider states where the subject sits (“in the left third, about 35\% across the frame”), how much of the frame it fills (“about 17\%”), and whether anything is cut off.

\textbf{360\textdegree{}~Panorama.} The 100 scenes were selected from the Poly Haven HDRI library with 30 urban, 25 indoor, 25 nature, 15 golden hour, and 5 night scenes. User preferences for 360\textdegree{}~framings demonstrate strong statistical significance. A Friedman test across the 100 evaluation scenes rejects equal mean ranks ($\chi^2 = 43.4$, $p < 10^{-9}$), while scene-level Wilcoxon signed-rank tests confirm significant preference for our framings over Venus ($p = 1.5 \times 10^{-8}$) and CACNet ($p = 1.6 \times 10^{-12}$), with identical conclusions holding at the participant level ($p < 10^{-16}$). Furthermore, scene-resampled bootstrapping estimates our rank-1 preference share at 53.2\% (95\% CI $[48.8\%, 57.6\%]$), substantially exceeding the chance level of 33.3\% across all resamples.

\section*{Reproducibility Statement}

\textbf{Code and assets.} Inference code for photographs and panoramas, the trained decision
layer, and the explainer adapter are released under Apache-2.0. Entry points are
one script per input type with \texttt{--mode fast|full} and \texttt{--explain}. Released
assets comprise the feature caches, the QLoRA adapter (SHA-256 recorded in the repository),
and the panoramic demo scenes.

\textbf{Determinism.} All models train with fixed seeds; \texttt{train.py --verify} retrains
the 13 decision-layer models and confirms byte-identical files. \texttt{reproduce.py}
recomputes the four results of Table\ref{tab:sota_comparison} from the released caches within tolerance
$5\times10^{-4}$.

\end{document}